\pdfoutput=1
\documentclass[11pt]{article}
\usepackage{easyReview}

\usepackage[]{EMNLP2023}

\usepackage[]{EMNLP2023}

\usepackage{svg}
\usepackage{enumitem}
\usepackage{times}
\usepackage{setspace}
\usepackage[normalem]{ulem}
\useunder{\uline}{\ul}{}
\usepackage{latexsym}
\usepackage{booktabs}
\usepackage[algo2e,linesnumbered,ruled,vlined]{algorithm2e} 
\usepackage{hyperref}
\usepackage{algpseudocode}
\usepackage[utf8]{inputenc}

\usepackage{microtype}

\usepackage{inconsolata}

\usepackage{tabularx}
\usepackage{graphicx}
\usepackage{amsthm}
\usepackage[T1]{fontenc}
\usepackage{amssymb}

\usepackage{ragged2e}
\usepackage{colortbl}
\usepackage[framemethod=TikZ]{mdframed}
\usepackage{subcaption}
\usepackage{amsmath} 
\usepackage{float}
\extrafloats{100}
\newif\ifcomments
\commentstrue

\newmdenv[%
    backgroundcolor=gray!10,
    linecolor=black,
    outerlinewidth=0.5pt,
    roundcorner=1mm,
    skipabove=\topsep,
    skipbelow=\topsep,
    font=\ttfamily\tiny,
]{promptbox}
\newcommand{\systemname}{\textsc{sPhinX}}
\newcommand{\mistral}{\textsc{Mistral-7B}}
\newcommand{\aya}{\textsc{Aya}}
\newcommand{\bactrian}{\textsc{Bactrian}}
\newcommand{\multialpaca}{\textsc{Multilingual Alpaca}}

\newcommand{\phismall}{\textsc{Phi-3-small}}

\title{\systemname:\ Sample Efficient Multilingual Instruction Fine-Tuning Through N-shot Guided Prompting}

\author{\parbox{0.9\linewidth}{\centering{Sanchit Ahuja$^\dag$\thanks{~~denotes equal contribution, $^\Delta$denotes equal advising, $^\diamondsuit$Work done when the authors were at Microsoft} \quad Kumar Tanmay$^{\heartsuit\diamondsuit}$\footnotemark[1] \quad Hardik Hansrajbhai Chauhan$^\dag$\\ 
\quad \textbf{Barun Patra$^\dag$} \quad \textbf{Kriti Aggarwal$^{\clubsuit\diamondsuit}$} \quad Luciano Del Corro$^\dag$ \\ 
\quad Arindam Mitra$^\dag$ \quad \textbf{Tejas Indulal Dhamecha$^{\beta\diamondsuit}$} \quad Ahmed Awadallah$^\dag$ \\
\quad \textbf{Monojit Choudhury$^{\spadesuit\diamondsuit}$} \quad \textbf{Vishrav Chaudhary$^{\alpha\diamondsuit\Delta}$} \quad \textbf{Sunayana Sitaram$^{\dag\Delta}$}\\
\vspace{2px}
{\rm $^\dag$Microsoft Corporation \quad $^\heartsuit$Harvard University \quad $^\beta$Adobe Research \\ \quad $^\spadesuit$MBZUAI University \quad $^\clubsuit$Hippocratic AI \quad $^\alpha$Meta AI} \\
\vspace{1px}
{\tt \{sanchitahuja205,kr.tanmay147\}@gmail.com}
}}}

\begin{document}
% \vskip{-10px}
\maketitle
\ \\

\vspace{10px}
% \ \\
% \vspace{-10px}
\begin{abstract}

Despite the remarkable success of large language models (LLMs) in English, a significant performance gap remains in non-English languages. To address this, we introduce a novel approach for strategically constructing a multilingual synthetic instruction tuning dataset, \systemname\ . Unlike prior methods that directly translate fixed instruction-response pairs, \systemname\ enhances diversity by selectively augmenting English instruction-response pairs with multilingual translations. Additionally, we propose \textit{LANGIT}, a novel N-shot guided fine-tuning strategy, which further enhances model performance by incorporating contextually relevant examples in each training sample. Our ablation study shows that our approach enhances the multilingual capabilities of \mistral\ and \phismall\, improving performance by an average of 39.8\% and 11.2\%, respectively, across multilingual benchmarks in reasoning, question answering, reading comprehension, and machine translation. Moreover, \systemname\ maintains strong performance on English LLM benchmarks while exhibiting minimal to no catastrophic forgetting, even when trained on 51 languages.

\end{abstract}

\section{Introduction}

Large Language Models (LLMs) have demonstrated exceptional performance across various tasks in English. However, their performance
in some non-English languages remains comparatively limited \cite{ahuja-etal-2023-mega, asai-etal-2024-buffet}. Further, the gap between the performance of Large Language Models (LLMs) and Small Language Models (SLMs) is more pronounced \citep{ahuja-etal-2024-megaverse} in non-English languages. \citet{Chinese-LLaMA-Alpaca} and \citet{balachandran2023tamil} utilize the method of fine-tuning models on datasets focused on particular languages. However, this can lead to catastrophic forgetting, which may negatively impact performance in English \cite{zhao2024llama, aggarwal2024exploring}.
Few techniques have been proposed to bridge this gap, such as incorporating better pre-training data in various languages and improving base tokenizers \cite{xu2024survey, dagan2024getting}. However, most of these changes need to be implemented in the pre-training stage, which demands extensive data and computational resources, making it practically unfeasible in many scenarios \cite{brown2020language}. Consequently, the most well-studied technique involves fine-tuning models for specific languages and tasks. Instruction fine-tuning (IFT) has become a popular technique to enhance the performance of language models in specific languages. This method combines the benefits of both the pre-training, fine-tuning, and prompting paradigms \cite{wei2021finetuned}.

Sample diversity is essential for effective instruction tuning in multilingual datasets. Many recent datasets have been generated by translating English content into other languages or by employing self-instruct techniques based on seed prompts \cite{li2023bactrianx, alpaca}. However, both methods can limit diversity. Machine translation may result in the loss of semantic nuance \citep{baroni2006new}, while self-instruct approaches often yield repetitive and homogeneous samples \cite{wang2022self}. This highlights the critical need for datasets that encompass a wide range of diverse samples.

% \citet{aggarwal2024maple} evaluate several models fine-tuned using Parameter Efficient fine-tuning and find that there is a gain in multilingual performance across several SLMs for many low-resource languages, with some high-resource languages performing worse after fine-tuning. However, the performance gains often do not match the performance of larger models, such as GPT-4 and Gemini and can be inconsistent across languages. Hence, there is a need to study instruction tuning for better multilingual performance in SLMs.  

In this paper, we present a novel recipe for creating a multilingual synthetic instruction tuning dataset, \systemname. \ It comprises 1.8M instruction-response pairs in 51 languages, derived by augmenting the Orca instruction tuning dataset samples\cite{mukherjee2023orca} through \textit{Selective Translated Augmentation} using GPT-4 \cite{achiam2023gpt}. We assess the effectiveness of \systemname\ by fine-tuning two models — \phismall\ and \mistral\ — across a range of evaluation benchmarks that test various language model capabilities across discriminative and generative tasks. We compare models fine-tuned on \systemname\ with those using other synthetic multilingual instruction tuning datasets like \aya\ \cite{ustun-etal-2024-aya}, \multialpaca\ \cite{alpaca}, and \bactrian\ \cite{li2023bactrianx} and observe significant performance gains across languages. We also compare our proposed translation strategy with translating the entire instruction using Azure Translator API, as is done with the popular multilingual synthetic IFT datasets to demonstrate the efficacy of our approach.
\\
The contributions of this paper are as follows:
\begin{itemize}
    \item We introduce a novel approach to generate synthetic data for multilingual instruction tuning by \textit{\textit{Selective Translated Augmentation}} of the Orca dataset with the assistance of GPT-4 (\S\ref{ssec:ds})
    
    % \item We conduct an in-depth analysis where we find \systemname\ to be more sample efficient and diverse across languages (\S\ref{ssec:data_analysis})
    \item We devise \textit{\textbf{LA}nguage-Specific \textbf{N}-shot \textbf{G}uided \textbf{I}nstruction \textbf{T}uning (LANGIT)} strategy for enhancing the multilingual capabilities of LLMs (\S\ref{ssec:fine-tuning_methodology})
    
    \item We also conduct extensive instruction tuning experiments on various multilingual instruction tuning datasets to evaluate generalizability in multilingual settings (\S\ref{sec:results}).

    \item We plan to release a subset of the augmented dataset by applying our strategy to the OpenOrca \footnote{\url{https://huggingface.co/datasets/Open-Orca/OpenOrca}} dataset \citep{OpenOrca} (\textsc{Open-sPhinX}) as well.
    
\end{itemize}

\section{Related Work}

\subsection{Multilingual Instruction fine-tuning} Early studies focused on fine-tuning pre-trained models on a variety of languages through data augmentation for a single task~\cite{hu2020xtreme,longpre2021mkqa,asai2022mia}. Currently, the approach has shifted to fine-tuning these models using a wide variety of tasks~\cite{10.5555/3618408.3619349, ouyang2022training}. Models such as BLOOMZ~\cite{muennighoff2022crosslingual} and mT0~\cite{muennighoff2022crosslingual} make significant strides in improving the multilingual performance of decoder-based models~\cite{ahuja-etal-2023-mega}. There have been multiple multilingual instruction datasets and models proposed such as Bactrian~\citep{li2023bactrianx}, \textsc{AYA}~\citep{ustun-etal-2024-aya}, \textsc{polyLM}~\citep{wei2023polylm} after BLOOMZ and mT0 \cite{muennighoff-etal-2023-crosslingual}. However, these models still do not perform as well as English in other languages, with the gap being huge for low-resource languages and languages written in scripts other than the Latin script ~\cite{ruder-etal-2021-xtreme,ahuja-etal-2023-mega,asai-etal-2024-buffet,ahuja-etal-2024-megaverse}. In this work, we aim to narrow the performance gap by introducing a strategy for creating datasets for multilingual instruction tuning and recipes for fine-tuning, which we will discuss in the following sections. 

\begin{figure*}[!ht]
    \centering
    \includegraphics[width=\textwidth]{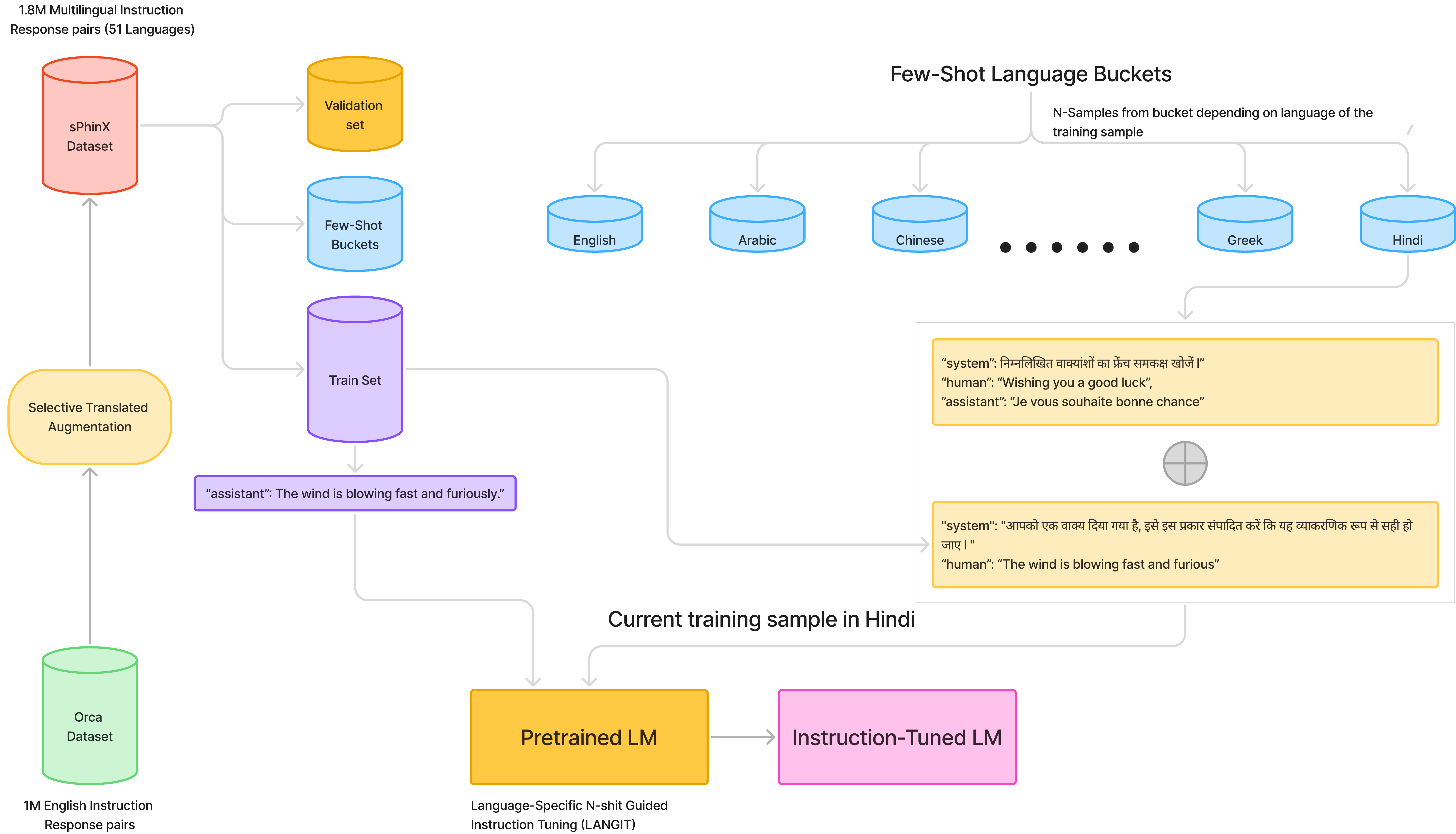}
    \caption{The figure above illustrates pipelines for \systemname\ data creation using Selective Translated Augmentation and Multilingual Instruction Tuning using \textit{LANGIT} strategy.}
    \label{fig:sphinx_pipeline}
\end{figure*}

\begin{figure*}[!ht]
    \centering
    \includegraphics[width=\textwidth]{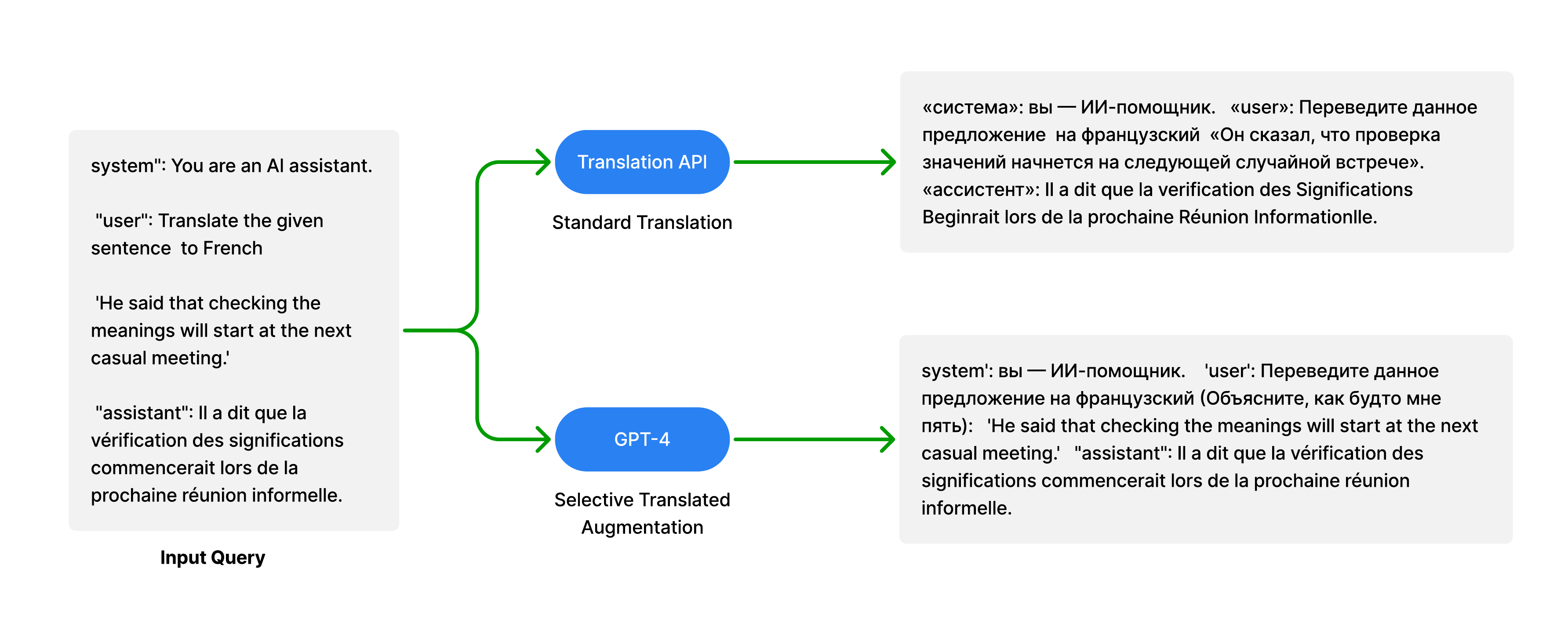}
    \caption{The figure compares the Translation API with Selective Translated Augmentation. The Translation API translates the entire input into Russian, while the Selective strategy localizes only necessary components. Here, system and user prompts are translated, but the input question and assistant’s response remain in the original language, preserving structure and intent.}
    \label{fig:prompt_fig}
\end{figure*}

\noindent
\subsection{Multilingual Synthetic Data Generation}
\vspace{-0.2em}
Most instruction-tuning datasets across multiple languages typically focus on general tasks rather than specific reasoning capabilities. Although datasets like Orca~\cite{mukherjee2023orca} and Orca 2~\cite{mitra2023orca} exist in English, they highlight a prevalent issue: current methods often prioritize style imitation over leveraging the reasoning abilities found in large foundation models (LFMs). The Orca dataset addresses this by imitating rich signals from GPT-4, including explanation traces and step-by-step thought processes~\cite{wei2023chainofthought}, guided by assistance from ChatGPT. In order to create multilingual datasets, researchers commonly use translation APIs or LLMs to translate English-specific datasets into target languages. For example, the Bactrian dataset~\cite{li2023bactrianx} translates Alpaca and Dolly instructions into 52 languages using the Google Translator API and generates outputs with GPT-3.5 turbo. Another example is of this work ~\cite{lai-etal-2024-llms}, which also utilizes Google Translation API for translating their source datasets. Our dataset approach aims to tackle these challenges by selectively translating only the essential portions of multilingual inputs. This strategy not only preserves semantic information but also accommodates diverse linguistic contexts, thereby enhancing the overall quality and applicability of instruction-tuning datasets across languages. In our work, we also show the pitfalls of training with data generated only using the Translation APIs such Google Translate or Azure Translate.

\section{ \systemname\ Dataset}
  In this section, we describe our dataset construction methodology (\S\ref{ssec:ds}), dataset filtering, and cleaning pipelines (\S\ref{sssec:filter}).

\subsection{Dataset Construction} \label{ssec:ds}

Inspired by \citep{mukherjee2023orca}'s work, we utilized the 1M GPT-4 generated instruction-response pairs from Orca and constructed our own dataset along similar lines using \textit{Selective Translated Augmentation} into 50 different languages with the help of GPT-4\footnote{GPT-4 inference hyper-parameters in Azure OpenAI interface set as: \texttt{temperature=0.0}}. We categorize them into three groups: high-resource, mid-resource, and low-resource languages as outlined in Table \ref{tab:resources}. For high-resource languages, we randomly sample 100k instruction-response pairs from the Orca 1M dataset and generate the responses from GPT-4 with \textit{Selective Translated Augmentation} as shown in Figure \ref{fig:prompt_fig}. Similarly, we leverage the same strategy for medium and low-resource languages by sampling 50k and 25k pairs respectively. Although GPT-4 performs competitively with commercial translation systems (Google Translate \& Bing Translate) it still lags on medium and low resource languages \cite{jiao2023chatgpt, hendy2023good}. Furthermore, as highlighted in \citep{chang2023multilinguality, lin2023speciality, 10.5555/3692070.3694291}, fine-tuning with a large set of samples from medium and low-resource languages might lead to catastrophic forgetting of high-resource languages. Therefore, we deliberately create fewer samples for medium and low-resource languages than for high-resource ones. Besides, \cite{shaham-etal-2024-multilingual} also demonstrates that a small number of multilingual training samples is sufficient to significantly boost multilingual performance, validating our approach of using fewer samples from medium- and low-resource languages.

A fundamental problem with using an off-the-shelf translation API is the lack of semantic and task awareness, in addition to \textit{translationese} \cite{baroni2006new}, which can result in poor quality training data. Consider for example the task of Machine Translation as part of the instruction, wherein the language of the source sentence should be retained. However, an off-the-shelf API, without task awareness, would translate it, resulting in an ambiguous instruction. To mitigate this issue, we used GPT-4 to augment the instructions using \textit{Selective Translated Augmentation}, so that task-specific components of instruction responses are translated into the appropriate language without changing the semantic meaning. Figure \ref{fig:example_translation} illustrates this with concrete examples. The first example demonstrates the aforementioned translation inconsistency issue for an instruction asking for a French equivalent of an English phrase. The second example demonstrates the consequence of direct translations in the \textsc{m-Alpaca} dataset: wherein the translation of the task input results in the task being ill-defined based on the instructions. As demonstrated, our proposed \textit{Selective Translated Augmentation} method is able to keep the semantic information of the task intact while translating the instructions. For the exact prompt, please refer to Figure \ref{fig:cautious prompt} in the Appendix.

\subsection{Dataset Filtering and Quality Assessment} \label{sssec:filter}

After creating the dataset, we filtered out samples where GPT-4 failed to generate a response. The final dataset comprised 1.8 million samples in 51 languages(Table: \ref{tab:lang_table}), divided into three subsets: Train, Validation, and Few-shot. Each language's dataset was partitioned to ensure that the Validation and Few-shot sets contained 2,000 and 1,000 samples, respectively, while the Train set included the remaining data. This approach guarantees consistent distributions across languages in the Validation and Few-shot sets, ensuring equitable representation regardless of the training distribution. The final split ratio for Train, Validation, and Few-shot sets was 92:5.3:2.7.

We also conducted a small-scale quality assessment of the generated data for languages such as Bengali, Hindi, German, Turkish, and Tamil. The researchers and engineers in our organization, who are native speakers of these languages, evaluated the data on the basis of coherence, fluency, and information retention. Our findings indicate that the generated dataset is moderate to high quality.

\subsection{Sample Diversity in \systemname} \label{sample_diversity}

Unlike prior multilingual datasets such as \bactrian\ and \textsc{m-Alpaca}, which translate a fixed set of instruction-response pairs into multiple languages, \systemname\ ensures diversity by sampling unique subsets of instruction-response pairs for each language.

For instance, \bactrian\ is constructed from 67k English instruction-response pairs (Alpaca + Dolly) and translated into 52 languages, resulting in identical samples across all languages. In contrast, \systemname\ samples from 1M GPT-4-generated instruction-response pairs, ensuring that no two languages share the exact same subset.

Mathematically, the probability that all samples in one language dataset \( A \) are also in another language dataset \( B \), when sampled without replacement from a larger dataset \( D \), is given by:
\noindent 
\[
P(\text{all A in B}) \approx \left(\frac{m}{N}\right)^n = \left(\frac{100{,}000}{1{,}000{,}000}\right)^{20{,}000}
\]
\vspace{-0.5em}

\noindent where:
\vspace{-0.5em}
\begin{itemize}[leftmargin=1.5em, itemsep=0pt, parsep=0pt]
    \item \( N = 1{,}000{,}000 \) (Total samples in \systemname),
    \item \( n = 20{,}000 \) (Samples in language \( A \)),
    \item \( m = 100{,}000 \) (Samples in language \( B \)).
\end{itemize}

Since the exponential term results in an extremely small probability, this confirms that no two languages have identical instruction-response sets in \systemname.

To further enhance diversity, we apply \textit{Selective Translated Augmentation}, translating 10\% of samples for high-resource languages, 5\% for mid-resource languages, and 2.5\% for low-resource languages. This ensures that translated content varies across languages, preventing uniformity.

Additionally, code-switching naturally emerges from this augmentation process, further increasing linguistic diversity. Compared to \aya, which exhibits moderate variation across task instructions, \systemname\ introduces greater sample diversity by leveraging a larger and more heterogeneous seed set \citep{mukherjee2023orca} and selective augmentation strategy. Exploring code-switching phenomena would be an interesting task in these synthetically generated datasets, but currently that is out-of-scope of for this work.

\section{\textit{LANGIT}} \label{ssec:fine-tuning_methodology}
Following Instruction Tuning strategies of ~\citep{10.5555/3618408.3619349} and also taking inspiration from \citep{min-etal-2022-metaicl}, we devise \textit{Language-Specific N-shot Guided Instruction fine-tuning (LANGIT)} Figure: \ref{fig:sphinx_pipeline}. This method aims to improve the model’s ability to follow instructions by augmenting training examples with additional context from a set of few-shot examples in the same language. This added context helps guide the model, enabling it to learn more effectively from the provided examples.

For each training example, we begin by sampling a number of few-shot examples, which are instruction-response pairs in the same language. The number of few-shot examples \(N\) is determined probabilistically, with a 30\% chance of selecting no few-shot examples, a 20\% chance of selecting one, and gradually lower probabilities for higher numbers of few-shots. The maximum number of few-shot examples we sample is six, due to constraints imposed by the model's context length (8192 tokens) and the typically higher tokenization length in languages other than English.

Once the number of few-shot examples is determined, they are prepended to the main training example, forming an augmented input. This augmented input is then fed into the model for instruction tuning. The purpose of this approach is to expose the model to additional examples of different tasks, helping it generalize better to new tasks in the same language.

We performed experiments to analyze how the model performs on each dataset when fine-tuned using the \textit{LANGIT} strategy detailed in the next section (\S\ref{sec:results}). Additionally, we fine-tuned the models on the \systemname\ dataset without using \textit{LANGIT} to provide a baseline for comparison. To assess the effectiveness of each instruction-tuning dataset on an equal scale, we conducted a comparative analysis of model performance on different benchmarks, fine-tuning each model on approximately 8 billion tokens per dataset using the \textit{LANG} strategy.

This fine-tuning strategy is consistently applied across datasets for both the \phismall\ and \mistral\ base models. A comparison of token lengths across different datasets is provided in Table \ref{tab:tokenlength}, showing the average token lengths as tokenized by the \phismall\ model.

\section{Experiments}

\subsection{Setup}
\textbf{Base Models}: We use \mistral\footnote{We specifically use the v1.0 base model from \\\url{https://huggingface.co/mistralai/Mistral-7B-v0.1}} and \phismall~\cite{abdin2024phi3} base model variants and instruction fine-tune them. \\ 

\noindent\textbf{Datasets}: Apart from the \systemname\ dataset, we use \textsc{Bactrian}~\cite{li2023bactrianx}, \textsc{m-Alpaca}~\cite{wei2023polylm} and \textsc{Aya}~\cite{singh2024aya} instruction datasets for comparative evaluation. We also utilize the Azure Translator API\footnote{\url{https://azure.microsoft.com/en-us/products/ai-services/ai-translator}} (\textsc{sPhinX-T}) to translate the original dataset into all our target languages, demonstrating the effectiveness of our \textit{Selective Translated Augmentation} approach. More details about the datasets used for comparative evaluation are present in Appendix \S\ref{ssec:baseline_datasets}. \\

\noindent \textbf{Evaluation}: We evaluate\footnote{Evaluation prompts and other details in Appendix \S\ref{ssec:app_prompts} and \S\ref{ssec:eval_benchmarks}.} our fine-tuned models along with the available base and Instruction fine-tuned model variants of \mistral\ and \phismall\ (IFT\footnote{We take the \mistral\ instruction-tuned variant from~\url{https://huggingface.co/mistralai/Mistral-7B-Instruct-v0.1} and \phismall\ variant from \url{https://huggingface.co/microsoft/Phi-3-small-8k-instruct}.}) on 4 discriminative tasks; XCOPA~\cite{ponti-etal-2020-xcopa}(4-shot), XStoryCloze~\cite{lin-etal-2022-shot}(4-shot), XWinograd~\cite{muennighoff-etal-2023-crosslingual}(0-shot),~\cite{tikhonov2021heads}, Belebele(0-shot)~\cite{bandarkar2023belebele}, and 2 generative tasks; XQuAD (3-shot)~\cite{artetxe-etal-2020-cross} and Translation (4-shot)~\cite{bojar-EtAl:2014:W14-33, bojar-EtAl:2016:WMT1, kocmi-etal-2023-findings} using the language model evaluation harness~\cite{eval-harness}. The number of few-shot selection are inspired from these works \cite{ahuja-etal-2023-mega, ahuja-etal-2024-megaverse, asai-etal-2024-buffet}

Apart from generative tasks such as XQuAD, and machine translation, we also evaluate our instruction-tuned models on open-ended generation prompts. For this, we use an LLM-based evaluation approach to simulate win rates. We use the open-source test set from the Aya Dataset \citep{singh-etal-2024-aya}, which includes 250 prompts per language across six languages. We use GPT-4o as the LLM evaluator to pick the preferred model generation on this test set, and we subsequently compute win rates (\%) based on these preferences. To avoid a potential
bias, we randomize the order of the models during the evaluation. The prompt for the evaluator is described in the Appendix: \S\ref{ssec:app_prompts}.

\section{Results} \label{sec:results}
\begin{figure}[!ht]
    \centering
    \includegraphics[width=0.51\textwidth]{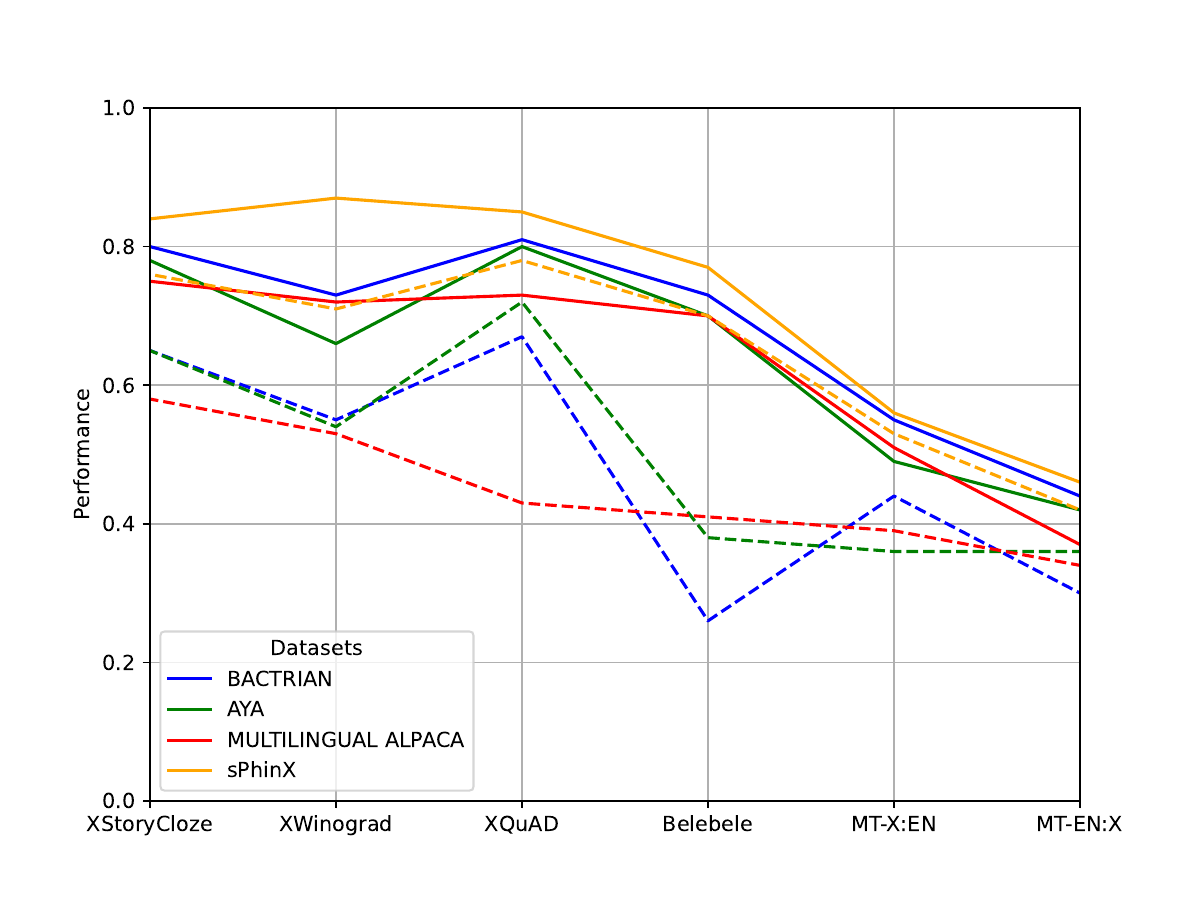}
    \caption{Performance of \mistral\ and \phismall\ when instruction-tuned on 8B tokens across various
datasets on different benchmarks. The solid lines represent the Phi fine-tuned models and the dashed lines represent the Mistral fine-tuned models.}
    \label{fig:sametokenperf}
\end{figure}

\begin{table*}
\centering
% \tiny
% \footnotesize
\footnotesize
\begin{tabular}{llllllll}
\toprule
\textbf{Model} & \multicolumn{1}{c}{$XC$} & \multicolumn{1}{c}{$XS$} & \multicolumn{1}{c}{$XW$} & \multicolumn{1}{c}{$XQ$} & \multicolumn{1}{c}{$BL$} & \multicolumn{1}{c}{$MT^1$} & {$MT^2$} \\ 
% \hline
\cmidrule(r){1-8}
\mistral &  &  &  &  &  &  &  \\ 
% \hline
\cmidrule(r){1-8}
Base Model & 0.63 & 0.68 & 0.52 & 0.74 & 0.24 & 0.54 & 0.42 \\
IFT & 0.62 & 0.73 & 0.54 & 0.60 & 0.47 & 0.49 & 0.39 \\
\textsc{m-Alpaca} & 0.55 & 0.59 & 0.51 & 0.46 & 0.41 & 0.41 & 0.39 \\
\aya & 0.68 & 0.71 & 0.54 & 0.66 & 0.38 & 0.39 & 0.37 \\
\bactrian & 0.54 & 0.67 & 0.54 & 0.69 & 0.26 & 0.45 & 0.34 \\
\systemname-T & 0.61 & 0.78 & 0.57 & 0.78 & 0.67 & 0.49 & 0.38 \\
\systemname-0s & 0.58 & 0.58 & 0.68 & 0.69 & 0.67 & 0.49 & 0.42 \\
\systemname & \textbf{0.68} & \textbf{0.81} & \textbf{0.71} & \textbf{0.80} & \textbf{0.71} & \textbf{0.55} & \textbf{0.46} \\ 
% \hline
\cmidrule(r){1-8}
\phismall &  &  &  &  &  &  &  \\ 
% \hline
\cmidrule(r){1-8}
Base Model & \multicolumn{1}{c}{0.64} & \multicolumn{1}{c}{0.78} & \multicolumn{1}{c}{0.75} & \multicolumn{1}{c}{0.78} & \multicolumn{1}{c}{0.65} & {0.54} & 0.42 \\
IFT & \multicolumn{1}{c}{0.68} & \multicolumn{1}{c}{0.79} & \multicolumn{1}{c}{0.78} & \multicolumn{1}{c}{0.75} & \multicolumn{1}{c}{0.70} & {0.54} & 0.46 \\
\textsc{m-Alpaca} & \multicolumn{1}{c}{0.68} & \multicolumn{1}{c}{0.79} & \multicolumn{1}{c}{0.81} & \multicolumn{1}{c}{0.77} & \multicolumn{1}{c}{0.75} & {0.45} & 0.39 \\
\aya & \multicolumn{1}{c}{0.65} & \multicolumn{1}{c}{0.79} & \multicolumn{1}{c}{0.69} & \multicolumn{1}{c}{0.83} & \multicolumn{1}{c}{0.72} & {0.41} & 0.40 \\
\bactrian & \multicolumn{1}{c}{0.71} & \multicolumn{1}{c}{0.82} & \multicolumn{1}{c}{0.73} & \multicolumn{1}{c}{0.85} & \multicolumn{1}{c}{0.77} & {0.54} & 0.40 \\
\systemname-T & \multicolumn{1}{c}{0.70} & \multicolumn{1}{c}{0.80} & \multicolumn{1}{c}{0.77} & \multicolumn{1}{c}{0.78} & \multicolumn{1}{c}{0.75} & {0.55} & 0.44 \\
\systemname-0s & \multicolumn{1}{c}{0.71} & \multicolumn{1}{c}{0.81} & \multicolumn{1}{c}{0.80} & \multicolumn{1}{c}{0.82} & \multicolumn{1}{c}{\uline{\textbf{0.79}}} & {\uline{\textbf{0.56}}} & 0.45 \\
\systemname & \multicolumn{1}{c}{\uline{\textbf{0.72}}} & \multicolumn{1}{c}{\uline{\textbf{0.84}}} & \multicolumn{1}{c}{\uline{\textbf{0.87}}} & \multicolumn{1}{c}{\uline{\textbf{0.87}}} & \multicolumn{1}{c}{\uline{\textbf{0.79}}} & {\uline{\textbf{0.56}}} & \textbf{\uline{0.46}} \\
\bottomrule
\end{tabular}
\caption{Performance of \mistral\ and \phismall\ instruction-tuned on various datasets. Abbreviations: $XC$ - XCOPA (Acc.,4-shot), $XS$ - XStoryCloze (Acc.,4-shot) , $XW$ - XWinograd (Acc., 0-shot), $XQ$ - XQuAD (F1,3-shot), $BL$ - Belebele (Acc., 0-shot). $MT^1$- Translation for x:en (ChrF, 4-shot), $MT^2$ - Translation for en:x direction (ChrF, 4-shot). The best performing dataset for each model is indicated in bold, and the overall best performing model is indicated with an underline.}
\label{tab:overall_avg_perf}
\end{table*}

\noindent We evaluate reasoning, question answering, translation and reading comprehension abilities of the \phismall\ and \mistral\ models, instruction-tuned on different multilingual datasets, using various benchmarks and find that fine-tuning on \systemname\ provides an average improvement of \textbf{39.8\%} and \textbf{11.2\%} respectively on both the models. (Refer to \textsc{sPhinX-0s} in Table \ref{tab:overall_avg_perf} for overall results and to Appendix \S\ref{ssec:all_tables} for language-wise results).
Additionally, as observed in Figure \ref{fig:sametokenperf}, the \systemname\ dataset significantly enhances the multilingual performance of the \phismall\ and \mistral\ model compared to other datasets even when fine-tuned on an equal number of tokens.

\section{Ablations}
\subsection{Improvements from \textit{LANGIT}}
To demonstrate the effectiveness of our \textit{LANGIT} strategy, we also instruction-tuned the models on \systemname\ with 0 shots, referring to this as \systemname-0s. As shown in Table \ref{tab:overall_avg_perf} (with detailed results in Appendix \S\ref{ssec:all_tables}), models fine-tuned on \systemname\ especially \mistral\ exhibit superior performance compared to its counterparts fine-tuned on other datasets across all benchmarks. Moreover, fine-tuning both \mistral\ and the \phismall\ on \systemname\ using the \textit{LANGIT} strategy further boosts the performance by an average of \textbf{15\%} and \textbf{3.2\%} respectively as compared to the vanilla fine-tuned model (\systemname\ -0s) across multilingual benchmarks.

Furthermore, employing the \textit{LANGIT} strategy leads to additional performance improvements indicating that \textit{LANGIT} can effectively enhance the multilingual capabilities of LLMs. From the detailed results in Appendix \S\ref{ssec:all_tables},  we observe no performance regression on high resource languages which normally occurs due to catastrophic forgetting \cite{chang2023multilinguality}. 

We also observe significant performance improvements in medium and low-resource languages such as Arabic, Hindi, Thai, Turkish, Tamil, and Telugu, further showcasing the effectiveness of our dataset and the \textit{LANGIT} fine-tuning strategy (Appendix \S\ref{ssec:all_tables}).

\subsection{Comparisons with the API translated dataset}

Due to the code-mixed nature of the instruction along with CoT reasoning explanations, a single sample of \systemname\ is notably richer as compared to its counterparts from the other datasets. This can be observed in the Table \ref{tab:overall_avg_perf} for \textsc{sPhinX-T} wherein the \systemname\ trained models with the \textit{LANGIT} strategy outperform the directly translated dataset baselines by an average of \textbf{11.7\%} and \textbf{6.3\%} for both \mistral\ and \phismall\ respectively when compared to the \textsc{sPhinX-T} baselines across multilingual benchmarks. Consequently, even with fewer samples as compared to the other datasets (keeping the number of the tokens the same), models trained on \systemname\ achieve better performance, thereby demonstrating the per-sample efficiency of \systemname.
\subsection{Simulated Preference Evaluation}
As shown in Table \ref{tab:win_rate}, our win-rate experiments reveal that the GPT-4o evaluator predominantly favored outputs generated by the Mistral base model trained on the \textsc{sPhinX} dataset using the \textit{LANGIT} strategy over other models. For the Phi baselines, we observed a higher percentage of TIEs for all the languages except English, where the evaluator rated both outputs equally, rather than favoring a specific model. This performance gap between the Mistral and Phi models likely arises from the age of their respective base models. Since the Mistral base model is older, it benefits more from additional training on our dataset, whereas the more recently released Phi models are already competitive enough on these benchmarks resulting in preferring both the outputs equally.

\begin{table*}[]
\centering
\scriptsize
\begin{tabular}{lcccccc}
\toprule
\textbf{Model}     & ar       & en       & po       & te       & tu       & zh       \\ \midrule
\mistral\ & \multicolumn{1}{l}{} & \multicolumn{1}{l}{} & \multicolumn{1}{l}{} & \multicolumn{1}{l}{} & \multicolumn{1}{l}{} & \multicolumn{1}{l}{} \\ \midrule
IFT                & \textbf{75} / 9 / 16  & 59 / \textbf{36} / 5  & 57 / 27 / \textbf{16} & 62 / 15 / \textbf{23} & \textbf{65} / 13 / 22 & \textbf{70} / 26 / 4  \\
\textsc{m-Alpaca}  & \textbf{85} / 2 / 13  & \textbf{74} / 21 / 5  & 52 / \textbf{33} / 15 & \textbf{69} / 8 / 23  & \textbf{70} / 4 / 25  & \textbf{75} / 15 / 10 \\
\textsc{Aya}       & \textbf{78} / 11 / 11 & \textbf{85} / 11 / 4  & 55 / 30 / \textbf{15} & 56 / 18 / \textbf{25} & 62 / \textbf{19} / 19 & \textbf{78} / 14 / 8  \\
\textsc{Bactrian}  & \textbf{82} / 4 / 13  & \textbf{85} / 12 / 3  & 57 / \textbf{31} / 12 & 56 / 18 / \textbf{26} & 62 / \textbf{20} / 18 & \textbf{74} / 14 / 12 \\
\textsc{sPhinX-T}  & 61 / \textbf{23} / 16 & 63 / \textbf{27} / 10 & 56 / 24 / \textbf{20} & 56 / 24 / \textbf{20} & 62 / \textbf{19} / 19 & 66 / \textbf{23} / 11 \\
\textsc{sPhinX-0s} & 65 / \textbf{19} / 16 & \textbf{74} / 20 / 5  & 55 / \textbf{31} / 14 & 51 / 22 / \textbf{27} & 52 / \textbf{36} / 12 & 68 / \textbf{20} / 11 \\ \midrule
\phismall\         &          &          &          &          &          &          \\ \midrule
IFT                & \textbf{46} / 38 / 17 & 55 / \textbf{43} / 2  & 50 / \textbf{44} / 6  & 39 / 29 / \textbf{36} & 30 / 27 / \textbf{43} & 50 / \textbf{44} / 6  \\
\textsc{m-Alpaca}  & \textbf{46} / 5 / 49  & 59 / \textbf{29} / 12 & 44 / 21 / \textbf{35} & 33 / 6 / \textbf{60}  & 31 / 10 / \textbf{59} & 30 / 6 / \textbf{64}  \\
\textsc{Aya}       & 40 / 18 / \textbf{42} & \textbf{80} / 11 / 9  & \textbf{56} / 16 / 28 & 37 / 18 / \textbf{46} & 25 / 21 / \textbf{54} & 36 / 22 / \textbf{41} \\
\textsc{Bactrian}  & 32 / 13 / \textbf{55} & \textbf{68} / 17 / 15 & 51 / 14 / \textbf{36} & 24 / 14 / \textbf{62} & 32 / 15 / \textbf{53} & 26 / 11 / \textbf{63} \\
\textsc{sPhinX-T}  & 35 / 14 / \textbf{51} & \textbf{75} / 12 / 13 & \textbf{61} / 12 / 27 & 23 / 14 / \textbf{63} & 36 / 18 / \textbf{47} & 37 / 12 / \textbf{51} \\
\textsc{sPhinX-0s} & 23 / 11 / \textbf{66} & 61 / 17 / \textbf{22} & 32 / 18 / \textbf{50} & 14 / 9 / \textbf{77}  & 15 / 10 / \textbf{74} & 14 / 7 / \textbf{79}  \\ \bottomrule
\end{tabular}
\caption{Win rates (\%) according to GPT-4o: The first value represents the percentage of outputs where the evaluator preferred the \textsc{sPhinX} and \textit{LANGIT} trained model. The second value indicates the percentage of outputs preferred from the target model. The third value reflects cases where the evaluator rated both outputs equally (TIE).}
\label{tab:win_rate}
\end{table*}

% \usepackage{booktabs}
\begin{table}[]
\scriptsize
\centering
\begin{tabular}{@{}ccc@{}}
\toprule
\multicolumn{1}{l}{\textbf{Benchmarks}}                                 & Base Model    & \systemname\  \\ \midrule
\begin{tabular}[c]{@{}c@{}}MMLU\\ \tiny(5-shot)\end{tabular}           & \textbf{0.76} & 0.75          \\
\begin{tabular}[c]{@{}c@{}}HellaSwag\\ \tiny(5-shot)\end{tabular}      & 0.81          & \textbf{0.83} \\
\begin{tabular}[c]{@{}c@{}}GSM-8k\\ \tiny(8-shot, CoT)\end{tabular}    & \textbf{0.85} & 0.77          \\
\begin{tabular}[c]{@{}c@{}}MedQA\\ \tiny(2-shot)\end{tabular}          & 0.64          & \textbf{0.66} \\
\begin{tabular}[c]{@{}c@{}}Arc-C\\ \tiny(10-shot)\end{tabular}         & \textbf{0.90} & \textbf{0.90} \\
\begin{tabular}[c]{@{}c@{}}Arc-E\\ \tiny(10-shot)\end{tabular}         & \textbf{0.97} & \textbf{0.97} \\
\begin{tabular}[c]{@{}c@{}}PIQA\\ \tiny(5-shot)\end{tabular}           & 0.84          & \textbf{0.89} \\
\begin{tabular}[c]{@{}c@{}}WinoGrande\\ \tiny(5-shot)\end{tabular}     & 0.77          & \textbf{0.82} \\
\begin{tabular}[c]{@{}c@{}}OpenBookQA\\ \tiny(10-shot)\end{tabular}    & 0.86          & \textbf{0.88} \\
\begin{tabular}[c]{@{}c@{}}BoolQ\\ \tiny(2-shot)\end{tabular}          & 0.82          & \textbf{0.87} \\
\begin{tabular}[c]{@{}c@{}}CommonSenseQA\\ \tiny(10-shot)\end{tabular} & 0.80          & \textbf{0.81} \\ \bottomrule
\end{tabular}
\caption{Performance of the \phismall\ base model and the \systemname\ tuned model on standard English LLM benchmarks.}
\label{tab:llm_benchmarks}
\end{table}

\subsection{Regression Analysis on Standard LLM Benchmarks}
It is well studied that training in multiple languages causes regression in performance in English due to catastrophic forgetting \cite{chang2023multilinguality}. We test this phenomenon for our trained models by checking the performance of the \phismall\ model fine-tuned with \systemname\ on English in the multilingual benchmarks we evaluate ((Appendix \S\ref{ssec:all_tables}) and on popular English-only benchmarks (Table \ref{tab:llm_benchmarks}).

We find that the \phismall\ fine-tuned on \systemname\ maintains its performance in English on the multilingual benchmarks and is also consistently able to maintain performance on standard English benchmarks such as MMLU (5-shot) \citet{hendrycks2021measuring}, MedQA (2-shot) \citet{jin2021disease}, Arc-C (10-shot), Arc-E (10-shot) \citet{clark2018think}, PiQA (5-shot) \citet{bisk2020piqa}, WinoGrande (5-shot) \citet{sakaguchi2021winogrande}, OpenBookQA (10-shot) \citet{mihaylov2018can}, BoolQ (2-shot) \citet{clark2019boolq} and CommonSenseQA (10-shot) \citet{talmor2018commonsenseqa} (Table \ref{tab:llm_benchmarks}). We notice some regression in the GSM-8k (8-shot, CoT) \citet{cobbe2021training} benchmark. This indicates that gains in multilingual performance caused by \systemname\ do not come at the cost of regression in English performance.
\section{Conclusion}
In this paper, we demonstrated how instruction tuning \mistral\ and \phismall\ on \systemname\ effectively improve their multilingual capabilities. We observed that instruction tuning the models using the \systemname\ dataset leads to performance improvement by an average of \textbf{39.8\%} and \textbf{11.2\%} for \mistral\ and \phismall\ respectively when compared to their corresponding base models across multilingual benchmarks.  Moreover, \systemname\ exhibits greater sample efficiency and diversity compared to other multilingual instruction tuning datasets. We also proposed \textit{LANGIT}, a strategy that enhances model performance by incorporating $N$ few-shot examples, boosting results by \textbf{15\%} and \textbf{3.2\%} for \mistral\ and \phismall, respectively, over vanilla fine-tuning with \systemname. Compared to the \textsc{sPhinX-T} translation baseline, \textit{LANGIT} yielded gains of \textbf{11.7\%} and \textbf{6.3\%}. Models fine-tuned on \systemname\ also showed improved performance in unseen languages without degrading English performance. We also observed that the GPT-4o win-rate evaluations favored \mistral\ with \textit{LANGIT}, while \phismall\ showed more ties due to its stronger baseline. Finally, we plan on releasing a subset of our augmented dataset built on OpenOrca (\textsc{Open-sPhinX}).

\section{Future Work}
All experiments were conducted using 7B base models with full fine-tuning. It would be interesting to explore our methods with adaptive fine-tuning techniques like LoRA \cite{hu2022lora} or PEFT \cite{peft}, and on smaller models, where we expect similar gains in multilingual performance. Our \textit{LANGIT} strategy uses $N$ examples from the same language and future work could investigate using $N$ examples from the same script to introduce greater diversity, especially for improving performance in low-resource languages. We also observed code-switched data
being generated when we employed this strategy to generate data. It will be interesting to explore this phenomena in a future study.

\section*{Limitations}
Our study has several limitations that can be considered in future research. Firstly, we conducted an extensive series of experiments, utilizing significant GPU resources and substantial time for model fine-tuning. Due to these resource-intensive processes, it may be difficult to apply our strategies to fully fine-tune a model. Besides, our study is confined to 7B models, explicitly excluding larger models. Despite this limitation, we believe that our methodologies are broadly applicable for fine-tuning smaller datasets using techniques like LoRA and PEFT. While some languages were not included, we made a conscious effort to cover a diverse set spanning multiple scripts and language families. 

\section*{Ethics Statement}
Despite our rigorous efforts to ensure that our dataset is free from discriminatory, biased, or false information, there remains a possibility that these problems are present, particularly in multilingual contexts. Hence, it is possible that these issues might propagate to our fine-tuned models as well. We are committed to mitigating such risks and strongly advocate for the responsible use of recipes and prevent any unintended negative consequences.

\bibliographystyle{acl_natbib}
\bibliography{custom}
\clearpage

\appendix

\section{Appendix}
\label{sec:appendix}

\subsection{Prompt Templates} \label{ssec:app_prompts}
Figure \ref{fig:cautious prompt} is the template for \textit{Selective Translated Augmentation} that was used to generate the synthetic data. Our reference dataset is in English and the \texttt{\{language\}} is the target language to generate the data in. Figure \ref{fig:xquad}, \ref{fig:xstorycloze}, \ref{fig:xwinograd},\ref{fig:xcopa}, \ref{fig:belebele} and \ref{fig:translation} are the prompts used to evaluate XQuAD, XstoryCloze, Xwinograd, XCOPA, Belebele and Translation respectively. Figure \ref{fig:win_rate_prompt} denotes the prompt used for simulating win-rate evaluations.

\subsection{Baseline Datasets For Comparative Evaluation} \label{ssec:baseline_datasets}
\begin{itemize}[itemsep=-2pt]
    \item \textsc{Bactrian} \cite{li2023bactrianx} is a machine translated dataset of the original alpaca-52k \cite{alpaca} and dolly-15k \cite{DatabricksBlog2023DollyV2} datasets into 52 languages. The instructions for this dataset were translated using a Translation API and then GPT-3.5-Turbo was prompted to generate outputs. We fine-tune our models on the complete dataset consisting of 3.4M instances.
    \item \textsc{m-Alpaca} \cite{wei2023polylm} is a self-instruct dataset that translates seed instructions from English to 11 languages, using GPT-3.5-Turbo for response generation. We fine-tune our models on the full dataset, which contains 500k data points.
    \item \textsc{Aya} \cite{singh-etal-2024-aya} contains human-curated prompt-completion pairs in 65 languages, along with 44 monolingual and multilingual instruction datasets and 19 translated datasets across 114 languages, totaling around 513M instances. To ensure parity with the \systemname\ dataset, we sampled it down to 2.7M instances, ensuring equal representation for each language in our subset.
\end{itemize}

\subsection{Evaluation Benchmarks} \label{ssec:eval_benchmarks}
\begin{itemize}[itemsep=-2pt]
    \item \textbf{XCOPA}: A causal commonsense reasoning dataset in 11 languages, evaluated in a 4-shot prompt setting.
    \item \textbf{XStoryCloze}: A professionally translated version of the English StoryCloze dataset~\cite{mostafazadeh2017lsdsem} in 10 languages, evaluated in a 4-shot prompt setting.
    \item \textbf{Belebele}: A parallel reading comprehension dataset across 122 languages, with evaluation on a subset of 14 languages in a 0-shot prompt setting.
    \item \textbf{XQuAD}: A QA dataset consisting of professional translations of a subset of SQuAD into 10 languages, evaluated in a 3-shot prompt setting due to context window limitations.
    \item \textbf{XWinograd}: A collection of Winograd Schemas in six languages for cross-lingual commonsense reasoning, evaluated in a 0-shot setting.
    \item \textbf{Translation}: We utilize a subset of WMT14, WMT16 and WMT23 of language pairs (7 languages), with evaluation in a 4-shot setting.
\end{itemize}

\subsection{Hyperparameters and Training Setup} \label{ssec:app_hyperparameters}
We used 5 nodes with each node containing 8 A100 GPUs with 80GB VRAM. These nodes communicated with each other using InfiniBand \footnote{\url{https://network.nvidia.com/pdf/whitepapers/IB_Intro_WP_190.pdf}}. We use DeepSpeed~\cite{10.1145/3394486.3406703} to do distributed fine-tuning over these GPUs. We use the same hyperparameters (Table \ref{tab:hyperparameters}) to fine-tune both \mistral\ and \phismall\ models.

\subsection{Detailed Results} 
\label{ssec:all_tables}
Tables \ref{tab:xwinograd}, \ref{tab:xquad}, \ref{tab:xcopa}, \ref{tab:xstorycloze}, \ref{tab:belebele}, \ref{tab:mt_x_en} and \ref{tab:mt_en_x}  show the granular results on our models and dataset.

\begin{figure}[!h]
\centering
\begin{promptbox}
\justify
\noindent Please carefully convert a conversation between a human and an AI assistant from English to {language}.  
The dialogue will be presented in JSON format, where 'system' denotes system instructions, 'human' indicates user queries, and 'assistant' refers to the AI's response.  
You should approach this task as if the 'human' original language is \{language\}.  
Translate the 'system' instructions fully into \{language\}. For the 'human' input, however, carefully discern which segments require translation into \{language\}, while leaving other parts in their original form.
 \\
For instance:
1. If the human contains a mix of languages, only translate the instruction part. \\
2. If the task is about language correction do not translate the target passage. \\

For the 'assistant' part, generate the 'assistant' response as you were prompted with ths newly translated system and assistant instructions.  
The outcome should retain the JSON format. Your response should solely contain the JSON.  
Do not translate the JSON keys.
\{"system": System text here, "human": User text here, "assistant": Assistant text here \}
\end{promptbox}
\caption{Prompt for Selective Translation using GPT-4}
\label{fig:cautious prompt}
\end{figure}

\begin{figure}[!h]
\centering
\begin{promptbox}
\justify
The task is to solve reading comprehension problems. You will be provided questions on a set of passages and you will need to provide the answer as it appears in the passage. The answer should be in the same language as the question and the passage. \\
Context: \\
\{context\} \\
Question: \\
\{question\} \\
Referring to the passage above, the correct answer to the given question is \{answer\}
\end{promptbox}
\caption{XQuAD evaluation prompt}
\label{fig:xquad}
\end{figure}

\begin{figure}[!h]
\centering
\begin{promptbox}
\justify
\{input\_sentence\_1\} \{input\_sentence\_2\} \\
\{input\_sentence\_3\} \{input\_sentence\_4\} \\
What is a possible continuation for the story given the following options? \\
Option1: \{sentence\_quiz1\}
Option1: \{sentence\_quiz2\}
% Option2: \{sentence_quiz2\} work here
\end{promptbox}
\caption{XstoryCloze evaluation prompt}
\label{fig:xstorycloze}
\end{figure}

\begin{figure}[!h]
\centering
\begin{promptbox}
\justify
Select the correct option out of option1 and option2 that will fill in the \_ in the below sentence: \\
\{sentence\} \\
Choices: \\
-option1: \{option1\} \\
-option2: \{option2\}
\end{promptbox}
\caption{Xwinograd evaluation prompt}
\label{fig:xwinograd}
\end{figure}

\begin{figure}[!h]
\centering
\begin{promptbox}
\justify
The task is to perform open-domain commonsense causal reasoning. You will be provided a premise and two alternatives, where the task is to select the alternative that more plausibly has a causal relation with the premise. Answer as concisely as possible in the same format as the examples below:
Given this premise: \\ 
\{premise\} \\ 
What's the best option? \\
-choice1 : \{choice1\} \\
-choice2 : \{choice2\} \\
We are looking for\{\% if question == \"cause\" \%\} a cause \{\% else \%\} an effect \{\% endif \%\}
\end{promptbox}
\caption{XCOPA evaluation prompt}
\label{fig:xcopa}
\end{figure}

\begin{figure}[!h]
\centering
\begin{promptbox}
\justify
The task is to perform reading comprehension task. Given the following passage, query, and answer choices, output only the letter corresponding to the correct answer. Do not give me any explanations to your answer. Just a single letter corresponding to the correct answer will suffice. \\ 
Passage: \{flores\_passage\} \\ 
Query: \{question\} \\
Choices: \\
A: \{mc\_answer1\} \\
B: \{mc\_answer2\} \\ 
C: \{mc\_answer3\} \\
D: \{mc\_answer4\}
\end{promptbox}
\caption{Belebele evaluation prompt}
\label{fig:belebele}
\end{figure}

\begin{figure}[!h]
\centering
\begin{promptbox}
\justify
Translate the following sentence pairs: \\ 
\{Source Language\}: \{Source Phrase\} \{Target Language\}: \{Target Phrase\}
\end{promptbox}
\caption{Translation evaluation prompt}
\label{fig:translation}
\end{figure}

\begin{figure}[!h]
\centering
\begin{promptbox}
\justify
\textbf{System}: You are a helpful following assistant whose goal is to select the preferred (least wrong) output for a given instruction in \{LANGUAGE\_NAME\}. \\
\textbf{User}: Which of the following answers is the best one for given instruction in \{LANGUAGE\_NAME\}.
A good answer should follow these rules:
1) It should be in \{[LANGUAGE\_NAME\}. \\
2) It should answer the request in the instruction. \\
3) It should be factually and semantically comprehensible. \\
4) It should be grammatically correct and fluent. \\
\\
Instruction: \{INSTRUCTION\} \\
Answer (A): \{COMPLETION A\} \\
Answer (B): \{COMPLETION B\} \\
\\
FIRST provide a one-sentence comparison of the two answers, explaining which you prefer and why. SECOND, on a new line, state only ‘Answer (A)’ or ‘Answer (B)’ to indicate your choice. If the both answers are equally good or bad, state ‘TIE’. Your response should use the format: \\
Comparison: <one-sentence comparison and explanation> \\
Preferred: <‘Answer (A)’ or ‘Answer (B)’ or ‘TIE’>
\end{promptbox}
\caption{Preference simulation prompt taken from \citep{ustun-etal-2024-aya} evaluation suite to evaluate our models on free-form generation using GPT-4o.}
\label{fig:win_rate_prompt}
\end{figure}

\begin{table}[h]
\centering
\begin{tabular}{@{}cc@{}}
\toprule
\multicolumn{1}{l}{\textbf{Hyperparameter}} & \multicolumn{1}{l}{\textbf{Value}} \\ \midrule
Batch Size                                  & 512                                \\
Context length                              & 8192                               \\
Learning Rate                               & $10^{-5}$                           \\
Scheduler                                   & Cosine                             \\
Epochs                                   & 10                             \\
Weight Decay                                   & 0.1                             \\

Optimizer                                   & AdamW                              \\ \bottomrule
\end{tabular}
\caption{Hyperparameters for model fine-tuning}
\label{tab:hyperparameters}
\end{table}
% \usepackage{booktabs}
% \usepackage{booktabs}
% \begin{table}[]
% \centering
% \begin{tabular}{@{}cc@{}}
% \toprule
% \multicolumn{1}{l}{\textbf{\# of shots}} & \multicolumn{1}{l}{\textbf{Weight}} \\ \midrule
% 0                                        & 0.3                                 \\
% 1                                        & 0.2                                 \\
% 2                                        & 0.1                                 \\
% 3                                        & 0.1                                 \\
% 4                                        & 0.1                                 \\
% 5                                        & 0.1                                 \\
% 6                                        & 0.1                                 \\ \bottomrule
% \end{tabular}
% \caption{The shot strategy and the corresponding weight assigned to the said strategy}
% \label{tab:weight_shots}
% \end{table}

\begin{table}[htbp]
\centering
\begin{tabular}{@{}
>{\columncolor[HTML]{FFFFFF}}l 
>{\columncolor[HTML]{FFFFFF}}l |
>{\columncolor[HTML]{FFFFFF}}l 
>{\columncolor[HTML]{FFFFFF}}l @{}}
\toprule
\(N\) & \(p(N)\) & \(N\) & \(p(N)\) \\ \midrule
0 & 0.3  & 4 & 0.1  \\
1 & 0.2  & 5 & 0.1  \\
2 & 0.1  & 6 & 0.1  \\
3 & 0.1  &   &      \\ \bottomrule
\end{tabular}
\caption{Probabilities of selecting number of shots in the \textit{LANG} strategy}
\label{tab:weight_shots}
\end{table}
% \usepackage{booktabs}
% \begin{table*}[htbp]
% \begin{tabular}{@{}
% >{\columncolor[HTML]{FFFFFF}}l 
% >{\columncolor[HTML]{FFFFFF}}r @{}}
% \toprule
% \textbf{Dataset} & \multicolumn{1}{l}{\cellcolor[HTML]{FFFFFF}Avg Token Length } \\ \midrule
% Aya                 & 2240 \\
% Bactrian            & 2465 \\
% Multilingual Alpaca & 1620 \\
% \systemname-0s             & 544 \\
% \systemname             & 3100 \\ \bottomrule
% \end{tabular}
% \caption{Average Token Length in each datasets}
% \label{tab:tokenlength}
% \end{table*}

\begin{table}[htbp]
  \centering
  \begin{tabular}{@{}>{\columncolor[HTML]{FFFFFF}}l 
                  >{\columncolor[HTML]{FFFFFF}}c@{}}
    \toprule
    \textbf{Dataset} & \multicolumn{1}{c}{Average Token} \\
    \textbf{ } & \multicolumn{1}{c}{Length/Sample} \\
    \midrule
    \aya                & 2240 \\
    \bactrian           & 2465 \\
    \textsc{m-Alpaca} & 1620 \\
    \textit{\systemname}-0s & 544 \\
    \textit{\systemname}    & 3100 \\
    \bottomrule
  \end{tabular}
  \caption{Average Token Length in each dataset}
  \label{tab:tokenlength}
\end{table}

\begin{table}[h]
\centering
\footnotesize
\begin{tabular}{@{}ll@{}}
\toprule
\textbf{High-Resource (100k)} &
  \begin{tabular}[c]{@{}l@{}}Spanish, Chinese Simplified, Japanese\\ French, German, Portuguese, Italian\end{tabular} \\ \midrule
\textbf{Mid-Resource (50k)} &
  \begin{tabular}[c]{@{}l@{}}Dutch, Swedish, Danish\\ Finnish, Russian, Norwegian\\ Korean, Chinese Traditional, Polish\\ Turkish, Arabic, Hebrew\\ Portuguese, Czech, Hungarian\end{tabular}
  \\ \midrule
\textbf{Low-Resource (25k)} &
  \begin{tabular}[c]{@{}l@{}}Indonesian, Thai, Greek\\ Slovak, Vietnamese, Slovenian\\ Croatian, Romanian, Lithuanian\\ Bulgarian, Serbian, Latvian\\ Ukranian, Estonian, Hindi\\ Burmese, Bengali, Afrikaan\\ Punjabi, Welsh, Icelandic\\ Marathi, Swahili, Nepali\\ Urdu, Telugu, Malayalam\\ Russian, Tamil, Oriya\end{tabular} \\ \bottomrule
\end{tabular}
\caption{Language distribution and samples across three tiers}
\label{tab:resources}
\end{table}

\begin{figure*}[!ht]
    \centering
    \includegraphics[width=\textwidth]{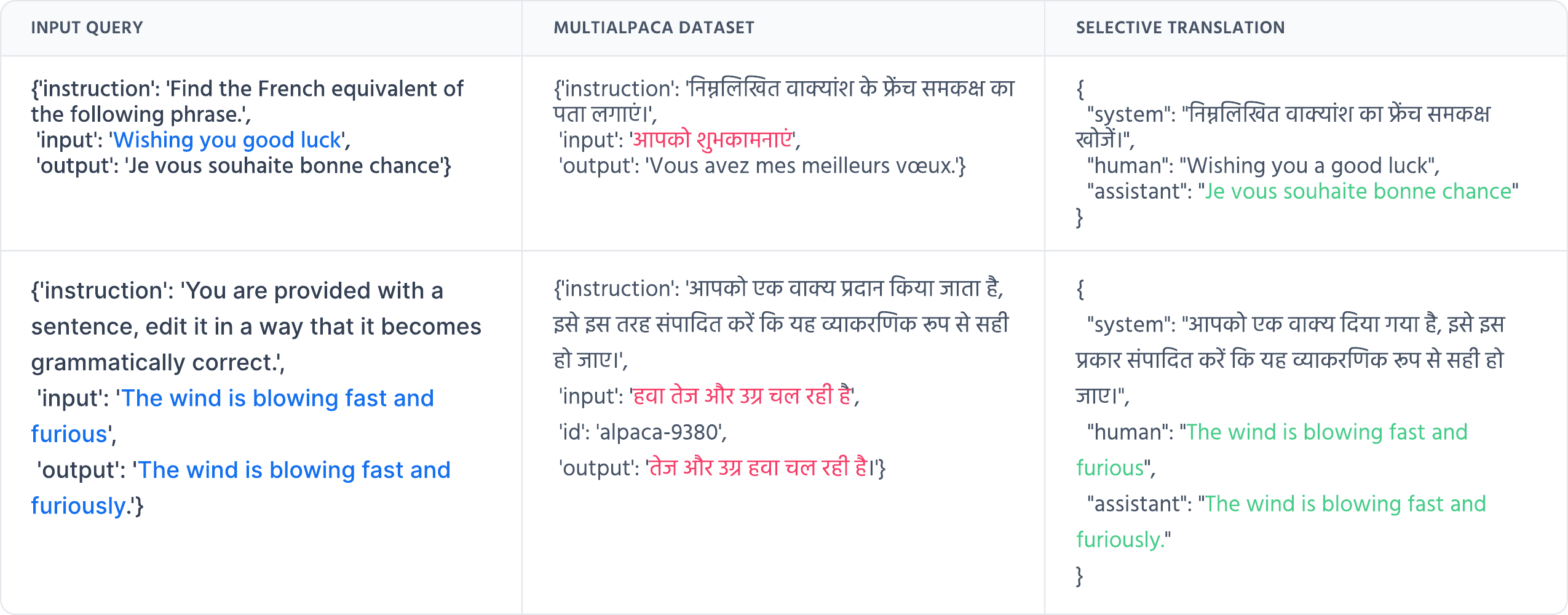}
    \caption{Some examples of input queries and its counterpart existing in the hindi version of the \textsc{Multialpaca} dataset and if it was generated using the Selective Translated Augmentation strategy. Again, we observe that the samples generated using Selective Translated Augmentation translate only the required amount of information as controlled via prompting whereas in \textsc{Multialpaca} the translations are direct translations where only a part of the instructions have been followed to translate the input queries.}
    \label{fig:example_translation}
\end{figure*}

\begin{table*}[]
\centering
\small
\begin{tabular}{cccccccc}
\cmidrule(r){1-8}
\textbf{Language} & en   & fr   & jp   & pt   & ru   & zh   & avg  \\ \cmidrule(r){1-8}
\mistral          &      &      &      &      &      &      &      \\ \cmidrule(r){1-8}
Base Model               & 0.52 & 0.47 & 0.52 & 0.54 & 0.54 & 0.50 & 0.52 \\
IFT               & 0.61 & 0.57 & 0.57 & 0.57 & 0.60 & 0.56 & 0.58 \\
\textsc{m-Alpaca}   & 0.61 & 0.57 & 0.57 & 0.57 & 0.60 & 0.56 & 0.58 \\
\aya              & 0.55 & 0.56 & 0.54 & 0.54 & 0.56 & 0.54 & 0.55 \\
\bactrian         & 0.61 & 0.57 & 0.57 & 0.57 & 0.60 & 0.56 & 0.58 \\
\systemname-T    & 0.58 & 0.61 & 0.57 & 0.53 & 0.54 & 0.57 & 0.57 \\
\systemname-0s    & 0.75 & 0.65 & 0.68 & 0.67 & 0.66 & 0.65 & 0.68 \\
\systemname\   & \textbf{0.80}       & \textbf{0.69}       & \textbf{0.72}       & \textbf{0.70}       & \textbf{0.67}       & \textbf{0.67}       & \textbf{0.71}       \\ \cmidrule(r){1-8}
\phismall         &      &      &      &      &      &      &      \\ \cmidrule(r){1-8}
Base Model               & 0.86 & 0.67 & 0.73& 0.77 & 0.74 & 0.72 & 0.75 \\
IFT               & 0.86 & 0.78 & 0.72 & 0.78 & 0.77 & 0.75 & 0.78 \\
\textsc{m-Alpaca}   & 0.87 & 0.76 & 0.75 & 0.78 & 0.76 & 0.71 & 0.81 \\
\aya              & 0.79 & 0.61 & 0.67 & 0.70 & 0.70 & 0.66 & 0.69 \\
\bactrian         & 0.83 & 0.72 & 0.71 & 0.75 & 0.70 & 0.68 & 0.73 \\
\systemname-T & 0.87                & 0.74                & 0.74              & 0.77         & 0.77                & 0.72                & 0.77                \\
\systemname-0s & 0.88                & 0.75                & 0.78                & {\ul \textbf{0.79}} & 0.81                & 0.76                & 0.80                \\
\systemname\   & {\ul \textbf{0.89}} & {\ul \textbf{0.76}} & {\ul \textbf{0.79}} & {\ul \textbf{0.79}}                & {\ul \textbf{0.82}} & {\ul \textbf{0.77}} & {\ul \textbf{0.84}} \\ \cmidrule(r){1-8}
\end{tabular}
\caption{Language-wise performance of instruction-tuned \mistral\ and \phismall\  models evaluated on XWinograd (0-shot). Metric: Accuracy. The best performing IFT dataset for each model is indicated in bold, and the overall best performing IFT model is indicated with an underline.}
\label{tab:xwinograd}
\end{table*}

\begin{table*}[]
\centering
\small
\begin{tabular}{cccccccccccccc}
\cmidrule(r){1-14}
\textbf{Language} &
  ar &
  de &
  el &
  en &
  es &
  hi &
  ro &
  ru &
  th &
  tr &
  vi &
  zh &
  avg \\ \cmidrule(r){1-14}
\mistral &
   &
   &
   &
   &
   &
   &
   &
   &
   &
   &
   &
   &
   \\ \cmidrule(r){1-14}
Base Model &
  0.62 &
  0.81 &
  0.64 &
  0.89 &
  0.86 &
  0.65 &
  0.82 &
  0.71 &
  0.59 &
  0.68 &
  0.79 &
  0.72 &
  0.73 \\
IFT &
  0.42 &
  0.68 &
  0.33 &
  0.92 &
  0.66 &
  0.5 &
  0.71 &
  0.61 &
  0.38 &
  0.63 &
  0.71 &
  0.68 &
  0.60 \\
\textsc{m-Alpaca} &
  0.10 &
  0.75 &
  0.15 &
  0.86 &
  0.82 &
  0.12 &
  0.62 &
  0.68 &
  0.12 &
  0.38 &
  0.52 &
  0.46 &
  0.46 \\
\aya &
  0.33 &
  0.73 &
  0.65 &
  0.85 &
  0.80 &
  0.63 &
  0.75 &
  0.67 &
  0.57 &
  0.61 &
  0.75 &
  0.59 &
  0.66 \\
\bactrian &
  0.67 &
  0.76 &
  0.26 &
  0.85 &
  0.86 &
  0.74 &
  0.77 &
  0.71 &
  0.59 &
  0.69 &
  0.77 &
  0.65 &
  0.69 \\
\systemname-T\ &
  0.71 &
  0.83 &
  0.75 &
  0.92 &
  0.89 &
  0.77 &
  0.84 &
  0.75 &
  0.60 &
  \textbf{0.77} &
  0.86 &
  0.63 &
  0.78 \\
\systemname-0s\ &
  0.54 &
  0.76 &
  0.70 &
  0.88 &
  0.84 &
  0.69 &
  0.77 &
  0.66 &
  0.52 &
  0.64 &
  0.71 &
  0.60 &
  0.69 \\
\systemname\ &
  \textbf{0.74} &
  \textbf{0.87} &
  \textbf{0.77} &
  \textbf{0.93} &
  \textbf{0.90} &
  \textbf{0.79} &
  \textbf{0.86} &
  \textbf{0.77} &
  \textbf{0.63} &
  \textbf{0.77} &
  \textbf{0.88} &
  {0.73} &
  \textbf{0.80} \\ \cmidrule(r){1-14}
\phismall &
   &
   &
   &
   &
   &
   &
   &
   &
   &
   &
   &
   &
   \\ \cmidrule(r){1-14}
   Base Model &
  0.68 &
  0.90 &
  0.77 &
  0.93 &
  0.91 &
  0.61 &
  0.84 &
  0.80 &
  0.55 &
  0.73 &
  0.86 &
  0.69 &
  0.78 \\
IFT &
  0.71 &
  0.88 &
  0.73 &
  0.92 &
  0.91 &
  0.64 &
  0.84 &
  0.80 &
  0.44 &
  0.70 &
  0.67 &
  0.76 &
  0.75 \\
\textsc{m-Alpaca} &
  0.55 &
  0.92 &
  0.74 &
  0.96 &
  0.94 &
  0.68 &
  0.87 &
  0.85 &
  0.50 &
  0.73 &
  0.88 &
  0.66 &
  0.77 \\
\aya &
  0.61 &
  0.89 &
  0.84 &
  0.94 &
  0.93 &
  0.80 &
  0.89 &
  0.82 &
  0.73 &
  0.83 &
  0.91 &
  0.79 &
  0.83 \\
\bactrian &
  0.81 &
  0.92 &
  0.81 &
  0.95 &
  0.95 &
  0.80 &
  0.90 &
  0.84 &
  0.72 &
  0.82 &
  0.91 &
  0.79 &
  0.85 \\
\systemname-T\ &
  0.80 &
  0.91 &
  0.82 &
  0.95 &
  0.94 &
  0.80 &
  0.90 &
  0.83 &
  0.69 &
  0.81 &
  0.91 &
  0.73 &
  0.78 \\
\systemname-0s\ &
  0.75 &
  0.89 &
  0.81 &
  0.94 &
  0.94 &
  0.75 &
  0.87 &
  0.79 &
  0.63 &
  0.77 &
  0.88 &
  0.78 &
  0.82 \\
\systemname\ &
    {\ul \textbf{0.84}} &
  {\ul \textbf{0.93}} &
  {\ul \textbf{0.87}} &
  {\ul \textbf{0.96}} &
  {\ul \textbf{0.96}} &
  {\ul \textbf{0.81}} &
  {\ul \textbf{0.91}} &
  {\ul \textbf{0.86}} &
  {\ul \textbf{0.73}} &
  {\ul \textbf{0.84}} &
  {\ul \textbf{0.92}} &
  {\ul \textbf{0.81}} &
  {\ul \textbf{0.87}} \\ \cmidrule(r){1-14}
\end{tabular}
\caption{Granular results for XQuAD (3-shot) on our model. Metric: F1. The best performing IFT dataset for each model is indicated in bold, and the overall best performing IFT model is indicated with an underline.}
\label{tab:xquad}
\end{table*}

\begin{table*}[]
\centering
\small
\begin{tabular}{cccccccccccccc}
\cmidrule(r){1-14}
\textbf{Language} & et   & ht   & id   & it   & qu   & sw   & ta   & th   & tr   & vi   & zh   & en   & avg  \\ \cmidrule(r){1-14}
\mistral          &      &      &      &      &      &      &      &      &      &      &      &      &      \\ \cmidrule(r){1-14}
Base Model               & 0.54 & 0.51 & 0.72 & 0.81 & 0.49 & 0.52 & 0.50 & 0.53 & 0.58 & 0.62 & 0.78 & 0.93 & 0.63 \\
IFT               & 0.52 & 0.52 & 0.69 & 0.79 & 0.50 & 0.51 & 0.50 & 0.54 & 0.57 & 0.63 & 0.75 & 0.90 & 0.62 \\
\textsc{m-Alpaca} & 0.51 & 0.50 & 0.52 & 0.63 & 0.50 & 0.50 & 0.50 & 0.51 & 0.51 & 0.49 & 0.65 & 0.74 & 0.55 \\
\aya              & 0.57 & 0.54 & 0.64 & 0.67 & 0.53 & 0.56 & 0.57 & 0.62 & 0.56 & 0.61 & 0.64 & 0.78 & 0.61 \\
\bactrian         & 0.52 & 0.50 & 0.53 & 0.60 & 0.49 & 0.51 & 0.50 & 0.51 & 0.51 & 0.52 & 0.52 & 0.71 & 0.54 \\
\systemname-T    & 0.57 & 0.50  & 0.64 & 0.72 & 0.50 & 0.50 & 0.58 & 0.57 & 0.57 & 0.62 & 0.71 & 0.83  & 0.61 \\
\systemname-0s    & 0.54 & 0.5  & 0.58 & 0.63 & 0.51 & 0.55 & 0.52 & 0.52 & 0.54 & 0.57 & 0.64 & 0.8  & 0.58 \\
\systemname\ &
  {\ul \textbf{0.64}} &
  \textbf{0.54} &
  \textbf{0.73} &
  \textbf{0.80} &
  \textbf{0.53} &
  {\ul \textbf{0.61}} &
  \textbf{0.59} &
  \textbf{0.63} &
  \textbf{0.67} &
  \textbf{0.66} &
  \textbf{0.80} &
  \textbf{0.91} &
  \textbf{0.68} \\ \cmidrule(r){1-14}
\phismall         &      &      &      &      &      &      &      &      &      &      &      &      &      \\ \cmidrule(r){1-14}
Base Model               & 0.55 & 0.51 & 0.80 & 0.93 & 0.52 & 0.54 & 0.46 & 0.56 & 0.61 & 0.66 & 0.86 & 0.98 & 0.64 \\
IFT               & 0.55 & 0.57 & 0.81 & 0.93 & 0.53 & 0.58 & 0.48 & 0.60 & 0.62 & 0.69 & 0.88 & 0.96 & 0.68 \\
\textsc{m-Alpaca} & 0.53 & 0.54 & 0.80 & 0.92 & 0.49 & 0.54 & 0.51 & 0.59 & 0.64 & 0.68 & 0.87 & 0.99 & 0.68 \\
\aya              & 0.60 & 0.55 & 0.72 & 0.83 & 0.52 & 0.55 & 0.52 & 0.62 & 0.59 & 0.69 & 0.75 & 0.89 & 0.65 \\
\bactrian         & \textbf{0.62} & 0.56 & 0.83 & 0.91 & 0.52 & \textbf{0.60} & 0.52 & 0.66 & 0.65 & 0.71 & 0.86 & 0.98 & 0.70 \\
\systemname-T    & 0.55 & 0.58 & 0.83 & 0.93 & 0.51 & 0.57 & 0.56 & 0.66 & 0.68 & 0.68 & 0.87 & 0.98 & 0.70 \\
\systemname-0s    & 0.59 & 0.58 & 0.84 & 0.93 & 0.50 & 0.60 & 0.54 & 0.63 & 0.68 & {\ul \textbf{0.72}} & 0.89 & 0.96 & 0.71 \\
\systemname\ &
  0.59 &
  {\ul \textbf{0.60}} &
  {\ul \textbf{0.85}} &
  {\ul \textbf{0.94}} &
  {\ul \textbf{0.52}} &
  {0.57} &
  {\ul \textbf{0.58}} &
  {\ul \textbf{0.68}} &
  {\ul \textbf{0.69}} &
  {{0.71}} &
  {\ul \textbf{0.90}} &
  {\ul \textbf{0.99}} &
  {\ul \textbf{0.72}} \\ \cmidrule(r){1-14}
\end{tabular}
\caption{Granular results for XCOPA (4-shot) on our model. Metric: Accuracy. The best performing IFT dataset for each model is indicated in bold, and the overall best performing IFT model is indicated with an underline.}
\label{tab:xcopa}
\end{table*}

\begin{table*}[]
\centering
\small
\begin{tabular}{ccccccccccccc}
\cmidrule(r){1-13}
\textbf{Language} & ar   & en     & es   & eu                  & hi   & id   & my   & ru   & sw            & te   & zh   & avg  \\ \cmidrule(r){1-13}
\mistral          &      &        &      &                     &      &      &      &      &               &      &      &      \\ \cmidrule(r){1-13}
Base Model               & 0.65 & 0.89   & 0.83 & 0.56                & 0.62 & 0.76 & 0.52 & 0.81 & 0.56          & 0.52 & 0.80 & 0.68 \\
IFT               & 0.70 & 0.95   & 0.92 & 0.54                & 0.69 & 0.79 & 0.57 & 0.90 & 0.58          & 0.54 & 0.88 & 0.73 \\
\textsc{m-Alpaca} & 0.53 & 0.73   & 0.70  & 0.51                & 0.51 & 0.57 & 0.50 & 0.66 & 0.52          & 0.52 & 0.71 & 0.59 \\
\aya              & 0.64 & 0.86   & 0.81 & 0.56                & 0.71 & 0.73 & 0.60 & 0.82 & 0.67          & 0.60 & 0.81 & 0.71 \\
\bactrian         & 0.69 & 0.82   & 0.74 & 0.52                & 0.59 & 0.76 & 0.54 & 0.73 & 0.62          & 0.61 & 0.76 & 0.67 \\
\systemname-T\   & 0.78 & 0.92   & 0.87 & 0.56                & 0.80 & 0.81 & 0.66 & 0.86 & 0.74          & 0.70 & 0.86 & 0.78 \\
\systemname-0s\   & 0.57 & 0.66   & 0.64 & 0.47                & 0.56 & 0.61 & 0.50 & 0.62 & 0.56          & 0.52 & 0.69 & 0.58 \\
\systemname\ &
  \textbf{0.83} &
  \textbf{0.96} &
  \textbf{0.94} &
  \textbf{0.57} &
  {\ul \textbf{0.84}} &
  \textbf{0.87} &
  {\ul \textbf{0.67}} &
  \textbf{0.91} &
  {\ul \textbf{0.80}} &
  {\ul \textbf{0.69}} &
  \textbf{0.94} &
  \textbf{0.81} \\ \cmidrule(r){1-13}
\phismall         &      &        &      &                     &      &      &      &      &               &      &      &      \\ \cmidrule(r){1-13}
Base Model               & 0.80 & 0.98   & 0.96 & 0.61                & 0.72 & 0.92 & 0.53 & 0.96 & 0.61          & 0.55 & 0.94 & 0.78 \\
IFT               & 0.81 & 0.98   & 0.96 & 0.61                & 0.75 & 0.92 & 0.56 & 0.96 & 0.61          & 0.53 & 0.94 & 0.79 \\
\textsc{m-Alpaca} & 0.81 & 0.98 & 0.98 & 0.58                & 0.76 & 0.93 & 0.52 & 0.97 & 0.64          & 0.54 & 0.96 & 0.79 \\
\aya              & 0.77 & 0.98   & 0.97 & 0.57                & 0.77 & 0.93 & 0.53 & 0.96 & 0.74          & 0.56 & 0.94 & 0.79 \\
\bactrian         & 0.83 & 0.98   & 0.98 & 0.61                & 0.83 & 0.94 & 0.54 & 0.97 & \textbf{0.79} & 0.63 & 0.94 & 0.82 \\
\systemname-T    & 0.84 & 0.98   & 0.98 & 0.60 & 0.80 & 0.95 & 0.52 & 0.96 & 0.72          & 0.60 & 0.85 & 0.80 \\
\systemname-0s    & 0.84 & 0.98   & 0.97 & {\ul \textbf{0.64}} & 0.77 & 0.95 & 0.52 & 0.96 & 0.74          & 0.57 & 0.95 & 0.81 \\
\systemname\ &
  {\ul \textbf{0.86}} &
  {\ul \textbf{0.99}} &
  {\ul \textbf{0.99}} &
  0.61 &
  \textbf{0.82} &
  {\ul \textbf{0.96}} &
  \textbf{0.54} &
  {\ul \textbf{0.98}} &
  0.74 &
  \textbf{0.61} &
  {\ul \textbf{0.97}} &
  {\ul \textbf{0.82}} \\ \cmidrule(r){1-13}
\end{tabular}
\caption{Granular results for XStoryCloze (4-shot) on our model. Metric: Accuracy. The best performing IFT dataset for each model is indicated in bold, and the overall best performing IFT model is indicated with an underline.}
\label{tab:xstorycloze}
\end{table*}

\begin{table*}[]
\footnotesize
\begin{tabular}{cccccccccccccccc}
\cmidrule(r){1-16}
\textbf{Language} &
  ar &
  de &
  es &
  en &
  fi &
  fr &
  hi &
  it &
  jp &
  ko &
  ta &
  te &
  vi &
  zh &
  avg \\ \cmidrule(r){1-16}
\mistral &
   &
   &
   &
   &
   &
   &
   &
   &
   &
   &
   &
   &
   &
   &
   \\ \cmidrule(r){1-16}
   Base Model &
  0.25 &
  0.23 &
  0.23 &
  0.24 &
  0.23 &
  0.23 &
  0.26 &
  0.24 &
  0.26 &
  0.23 &
  0.23 &
  0.25 &
  0.26 &
  0.25 &
  0.24 \\
IFT &
  0.32 &
  0.60 &
  0.62 &
  0.74 &
  0.36 &
  0.62 &
  0.32 &
  0.61 &
  0.43 &
  0.47 &
  0.27 &
  0.27 &
  0.39 &
  0.58 &
  0.47 \\
\textsc{m-Alpaca} &
  0.32 &
  0.50 &
  0.53 &
  0.56 &
  0.45 &
  0.51 &
  0.27 &
  0.51 &
  0.40 &
  0.41 &
  0.26 &
  0.26 &
  0.33 &
  0.48 &
  0.41 \\
\aya &
  0.34 &
  0.43 &
  0.43 &
  0.48 &
  0.38 &
  0.47 &
  0.35 &
  0.44 &
  0.4 &
  0.36 &
  0.27 &
  0.25 &
  0.37 &
  0.42 &
  0.38 \\
\bactrian &
  0.24 &
  0.27 &
  0.25 &
  0.25 &
  0.26 &
  0.27 &
  0.24 &
  0.28 &
  0.26 &
  0.26 &
  0.23 &
  0.23 &
  0.34 &
  0.28 &
  0.26 \\
\systemname-T &
  0.66 &
  0.74 &
  0.71 &
  0.81 &
  0.66 &
  0.76 &
  0.57 &
  0.73 &
  0.66 &
  0.68 &
  0.54 &
  0.46 &
  0.66 &
  0.71 &
  0.68 \\  
\systemname-0s &
  0.64 &
  0.75 &
  \textbf{0.75} &
  0.82 &
  0.66 &
  0.79 &
  0.53 &
  0.73 &
  0.69 &
  0.66 &
  0.48 &
  0.44 &
  0.66 &
  0.75 &
  0.67 \\
\systemname\ &
  \textbf{0.69} &
  \textbf{0.80} &
  0.69 &
  \textbf{0.87} &
  \textbf{0.71} &
  \textbf{0.82} &
  {\ul \textbf{0.60}} &
  \textbf{0.79} &
  \textbf{0.73} &
  \textbf{0.73} &
  {\ul \textbf{0.56}} &
  {\ul \textbf{0.48}} &
  \textbf{0.70} &
  \textbf{0.80} &
  \textbf{0.71} \\ \cmidrule(r){1-16}
\phismall &
   &
   &
   &
   &
   &
   &
   &
   &
   &
   &
   &
   &
   &
   &
   \\ \cmidrule(r){1-16}
Base Model &
  0.54 &
  0.87 &
  0.85 &
  0.92 &
  0.58 &
  0.86 &
  0.41 &
  0.86 &
  0.70 &
  0.58 &
  0.26 &
  0.30 &
  0.62 &
  0.82 &
  0.65 \\
IFT &
  0.63 &
  0.89 &
  0.88 &
  0.93 &
  0.63 &
  0.88 &
  0.48 &
  0.88 &
  0.77 &
  0.68 &
  0.32 &
  0.32 &
  0.68 &
  0.85 &
  0.70 \\
\textsc{m-Alpaca} &
  0.65 &
  0.92 &
  0.90 &
  0.94 &
  0.74 &
  0.91 &
  0.54 &
  0.90 &
  0.80 &
  0.70 &
  0.47 &
  \textbf{0.45} &
  0.72 &
  0.84 &
  0.75 \\
Aya &
  0.58 &
  0.86 &
  0.85 &
  0.91 &
  0.65 &
  0.87 &
  0.50 &
  0.86 &
  0.76 &
  0.67 &
  0.37 &
  0.35 &
  0.69 &
  0.84 &
  0.70 \\
\bactrian &
  0.67 &
  0.88 &
  0.88 &
  0.92 &
  0.70 &
  0.88 &
  0.51 &
  0.86 &
  0.77 &
  0.70 &
  0.37 &
  0.37 &
  0.74 &
  0.86 &
  0.72 \\
\systemname-T &
  0.71 &
  0.90 &
  0.90 &
  0.93 &
  0.73 &
  0.90 &
  0.56 &
  0.89 &
  0.82 &
  0.75 &
  0.42 &
  0.38 &
  0.75 &
  0.87 &
  0.75 \\
\systemname-0s &
  0.73 &
  0.91 &
  0.90 &
  0.93 &
  0.75 &
  0.92 &
  0.57 &
  0.91 &
  0.82 &
  {\ul \textbf{0.82}} &
  0.45 &
  {{0.40}} &
  0.76 &
  {\textbf{\uline{0.89}}} &
  0.77 \\
\systemname\ &
  {\ul \textbf{0.74}} &
  {\ul \textbf{0.93}} &
  {\ul \textbf{0.91}} &
  {\ul \textbf{0.94}} &
  {\ul \textbf{0.77}} &
  {\ul \textbf{0.93}} &
  {\textbf{0.58}} &
  {\ul \textbf{0.92}} &
  {\ul \textbf{0.84}} &
  0.76 &
  {\ul \textbf{0.46}} &
  {{0.40}} &
  {\ul \textbf{0.78}} &
  {\ul \textbf{0.89}} &
  {\ul \textbf{0.79}} \\ \cmidrule(r){1-16}
\end{tabular}
\caption{Granular results for Belebele (0-shot) on our model. Metric: Accuracy. The best performing IFT dataset for each model is indicated in bold, and the overall best performing IFT model is indicated with an underline.}
\label{tab:belebele}
\end{table*}

\begin{table*}
\centering
\small
\label{tab:overall_perf}
\begin{tabular}{lllllllll} 
\toprule
\textbf{Language} & \multicolumn{1}{c}{ar} & \multicolumn{1}{c}{fr} & \multicolumn{1}{c}{de} & \multicolumn{1}{c}{ro} & \multicolumn{1}{c}{ja} & \multicolumn{1}{c}{ru} & zh & avg \\ 
% \hline
\cmidrule(r){1-9}
\mistral &  &  &  &  &  &  &  &  \\ 
% \hline
\cmidrule(r){1-9}
Base Model & 0.48 & 0.63 & 0.65 & 0.56 & 0.43 & 0.54 & 0.48 & 0.54 \\
IFT & 0.37 & 0.61 & 0.61 & 0.58 & 0.28 & 0.5 & 0.49 & 0.49 \\
\textsc{m-Alpaca} & 0.34 & 0.51 & 0.51 & 0.46 & 0.29 & 0.36 & 0.41 & 0.41 \\
\aya & 0.30 & 0.50 & 0.51 & 0.46 & 0.24 & 0.35 & 0.41 & 0.39 \\
\bactrian & 0.40 & 0.55 & 0.52 & 0.51 & 0.34 & 0.44 & 0.42 & 0.45 \\
\systemname-T & 0.39 & 0.60 & 0.60 & 0.54 & 0.39 & 0.49 & 0.48 & 0.49 \\
\systemname-0s & 0.40 & 0.57 & 0.61 & 0.53 & 0.40 & 0.51 & 0.44 & 0.49 \\
\systemname & \textbf{0.45} & \textbf{0.63} & \textbf{0.64} & \textbf{0.59} & \textbf{0.45} & \textbf{\uline{0.55}} & \textbf{0.52} & \textbf{0.54} \\ 
% \hline
\cmidrule(r){1-9}
\phismall &  &  &  &  &  &  &  &  \\ 
% \hline
\cmidrule(r){1-9}
Base Model & \multicolumn{1}{c}{0.48} & \multicolumn{1}{c}{0.61} & \multicolumn{1}{c}{0.65} & \multicolumn{1}{c}{0.57} & \multicolumn{1}{c}{0.45} & \multicolumn{1}{c}{0.52} & 0.53 & 0.54 \\
IFT & \multicolumn{1}{c}{0.43} & \multicolumn{1}{c}{0.62} & \multicolumn{1}{c}{0.63} & \multicolumn{1}{c}{0.52} & \multicolumn{1}{c}{0.41} & \multicolumn{1}{c}{0.49} &  0.50 & 0.54   \\
\textsc{m-Alpaca} & \multicolumn{1}{c}{0.44} & \multicolumn{1}{c}{0.60} & \multicolumn{1}{c}{0.61} & \multicolumn{1}{c}{0.56} & \multicolumn{1}{c}{0.32} & \multicolumn{1}{c}{0.44} & 0.19 & 0.45 \\
\aya & \multicolumn{1}{c}{0.44} & \multicolumn{1}{c}{0.56} & \multicolumn{1}{c}{0.57} & \multicolumn{1}{c}{0.54} & \multicolumn{1}{c}{0.12} & \multicolumn{1}{c}{0.45} & 0.17 & 0.41 \\
\bactrian & \multicolumn{1}{c}{0.48} & \multicolumn{1}{c}{0.63} & \multicolumn{1}{c}{0.65} & \multicolumn{1}{c}{0.59} & \multicolumn{1}{c}{0.45} & \multicolumn{1}{c}{0.52} & 0.52 & 0.54 \\
\systemname-T & \multicolumn{1}{c}{0.48} & \multicolumn{1}{c}{0.64} & \multicolumn{1}{c}{0.65} & \multicolumn{1}{c}{0.59} & \multicolumn{1}{c}{0.46} & \multicolumn{1}{c}{0.53} & 0.52 & 0.55 \\
\systemname-0s & \multicolumn{1}{c}{\textbf{\uline{0.49}}} & \multicolumn{1}{c}{0.63} & \multicolumn{1}{c}{0.65} & \multicolumn{1}{c}{0.59} & \multicolumn{1}{c}{\textbf{\uline{0.46}}} & \multicolumn{1}{c}{\textbf{0.54}} & \textbf{\uline{0.53}} & \textbf{\uline{0.56}} \\
\systemname & \multicolumn{1}{c}{\textbf{\uline{0.49}}} & \multicolumn{1}{c}{\textbf{\uline{0.64}}} & \multicolumn{1}{c}{\textbf{\uline{0.66}}} & \multicolumn{1}{c}{\textbf{\uline{0.60}}} & \multicolumn{1}{c}{\textbf{\uline{0.46}}} & \multicolumn{1}{c}{\textbf{0.54}} & \textbf{\uline{0.53}} & \textbf{\uline{0.56}} \\
\bottomrule
\end{tabular}
\caption{Granular results for Translation for language to English direction (4-shot) on our model. Metric: ChrF. The best performing dataset for each model is indicated in bold, and the overall best performing model is indicated with an underline.}
\label{tab:mt_x_en}
\end{table*}

\begin{table*}
\centering
\small
\label{tab:overall_perf}
\begin{tabular}{lllllllll} 
\toprule
\textbf{Language} & \multicolumn{1}{c}{ar} & \multicolumn{1}{c}{fr} & \multicolumn{1}{c}{de} & \multicolumn{1}{c}{ro} & \multicolumn{1}{c}{ja} & \multicolumn{1}{c}{ru} & zh & avg \\ 
% \hline
\cmidrule(r){1-9}
\mistral &  &  &  &  &  &  &  &  \\ 
% \hline
\cmidrule(r){1-9}
Base Model & 0.29 & 0.60 & 0.54 & 0.52 & 0.21 & 0.34 & 0.47 & 0.42 \\
IFT & 0.16 & 0.58 & 0.57 & 0.48 & 0.17 & 0.29 & 0.43 & 0.38 \\
\textsc{m-Alpaca} & 0.15 & 0.62 & 0.66 & 0.40 & 0.12 & 0.48 & 0.44 & 0.41 \\
\aya & 0.12 & 0.54 & 0.63 & 0.48 & 0.16 & \textbf{\uline{0.39}} & 0.42 & 0.39 \\
\bactrian & 0.14 & 0.51 & 0.49 & 0.44 & 0.10 & 0.35 & 0.40 & 0.38 \\
\systemname-T & 0.31 & 0.58 & 0.53 & 0.47 & 0.20 & 0.30 & 0.26 & 0.38 \\
\systemname-0s & 0.30 & 0.60 & 0.55 & 0.51 & 0.22 & 0.31 & 0.45 & 0.42 \\
\systemname & \textbf{0.35} & \textbf{0.61} & \textbf{0.60} & \textbf{\uline{0.55}} & \textbf{0.26} & 0.36 & \textbf{\uline{0.49}} & \textbf{0.46} \\ 
% \hline
\cmidrule(r){1-9}
\phismall &  &  &  &  &  &  &  &  \\ 
% \hline
\cmidrule(r){1-9}
Base Model & \multicolumn{1}{c}{0.31} & \multicolumn{1}{c}{0.63} & \multicolumn{1}{c}{0.60} & \multicolumn{1}{c}{0.43} & \multicolumn{1}{c}{0.24} & \multicolumn{1}{c}{0.30} & 0.45 & 0.42 \\
IFT & \multicolumn{1}{c}{0.29} & \multicolumn{1}{c}{0.61} & \multicolumn{1}{c}{0.58} & \multicolumn{1}{c}{0.39} & \multicolumn{1}{c}{0.21} & \multicolumn{1}{c}{0.27} &  0.41 & 0.46   \\
\textsc{m-Alpaca} & \multicolumn{1}{c}{0.39} & \multicolumn{1}{c}{0.60} & \multicolumn{1}{c}{0.58} & \multicolumn{1}{c}{0.31} & \multicolumn{1}{c}{0.11} & \multicolumn{1}{c}{0.24} & 0.47 & 0.38 \\
\aya & \multicolumn{1}{c}{0.17} & \multicolumn{1}{c}{0.62} & \multicolumn{1}{c}{0.56} & \multicolumn{1}{c}{0.45} & \multicolumn{1}{c}{0.22} & \multicolumn{1}{c}{0.28} & 0.46 & 0.39 \\
\bactrian & \multicolumn{1}{c}{0.30} & \multicolumn{1}{c}{0.60} & \multicolumn{1}{c}{0.56} & \multicolumn{1}{c}{0.47} & \multicolumn{1}{c}{0.18} & \multicolumn{1}{c}{0.24} & 0.44 & 0.40 \\
\systemname-T & \multicolumn{1}{c}{0.33} & \multicolumn{1}{c}{0.63} & \multicolumn{1}{c}{0.60} & \multicolumn{1}{c}{0.48} & \multicolumn{1}{c}{0.24} & \multicolumn{1}{c}{0.31} & 0.48 & 0.43 \\
\systemname-0s & \multicolumn{1}{c}{0.32} & \multicolumn{1}{c}{0.63} & \multicolumn{1}{c}{\textbf{\uline{0.61}}} & \multicolumn{1}{c}{0.50} & \multicolumn{1}{c}{\textbf{\uline{0.27}}} & \multicolumn{1}{c}{\textbf{0.36}} & \textbf{\uline{0.49}} & 0.45 \\
\systemname & \multicolumn{1}{c}{\textbf{\uline{0.35}}} & \multicolumn{1}{c}{\textbf{\uline{0.64}}} & \multicolumn{1}{c}{\textbf{\uline{0.61}}} & \multicolumn{1}{c}{\textbf{0.51}} & \multicolumn{1}{c}{0.26} & \multicolumn{1}{c}{0.33} & \textbf{\uline{0.49}} & \textbf{\uline{0.46}} \\
\bottomrule
\end{tabular}
\caption{Granular results for Translation for English to language direction (4-shot) on our model. Metric: ChrF. The best performing dataset for each model is indicated in bold, and the overall best performing model is indicated with an underline.}
\label{tab:mt_en_x}
\end{table*}
% \input{emnlp2023-latex/tables/overall_average_perf}

% % Please add the following required packages to your document preamble:
% % \usepackage{booktabs}
% \begin{table*}[h]
% \centering
% \begin{tabular}{@{}ll@{}}
% \toprule
% \textbf{Tier 1 (100k)} &
%   \begin{tabular}[c]{@{}l@{}}Spanish, Chinese Simplified, Japanese\\ French, German, Portuguese, Italian\end{tabular} \\ \midrule
% \textbf{Tier 2 (50k)} &
%   \begin{tabular}[c]{@{}l@{}}Dutch, Swedish, Danish\\ Finnish, Russian, Norwegian\\ Korean, Chinese Traditional, Polish\\ Turkish, Arabic, Hebrew\\ Portuguese, Czech, Hungarian\end{tabular}
%   \\ \midrule
% \textbf{Tier 3 (25k)} &
%   \begin{tabular}[c]{@{}l@{}}Indonesian, Thai, Greek\\ Slovak, Vietnamese, Slovenian\\ Croatian, Romanian, Lithuanian\\ Bulgarian, Serbian, Latvian\\ Ukranian, Estonian, Hindi\\ Burmese, Bengali, Afrikaan\\ Punjabi, Welsh, Icelandic\\ Marathi, Swahili, Nepali\\ Urdu, Telugu, Malayalam\\ Russian, Tamil, Oriya\end{tabular} \\ \bottomrule
% \end{tabular}
% \caption{Language distribution and samples across three tiers}
% \label{tab:resource}
% \end{table*}

% \usepackage{booktabs}
% \usepackage[normalem]{ulem}
\begin{table*}[h]
\centering
\small
\begin{tabular}{@{}ccccc@{}}
\toprule
Code    & Languages            & Script       & Data   \\ \midrule
af      & Afrikaan             & Latin        & 20206  \\
ar      & Arabic               & Arabic       & 26803  \\
bn      & Bengali              & Bengal       & 20165  \\
bg      & Bulgarian            & Cyrillic     & 17300  \\
my      & Burmese              & Burmese      & 12123  \\
zh-Hans & Chinese\_Simplified  & Han          & 100650 \\
zh-Hant & Chinese\_Traditional & Hant         & 32363  \\
hr      & Croatian             & Latin        & 17340  \\
cs      & Czech                & Latin        & 32711  \\
da      & Danish               & Latin        & 36348  \\
nl      & Dutch                & Latin        & 36586  \\
en      & English              & Latin        & 199900 \\
et      & Estonian             & Latin        & 17207  \\
fi      & Finnish              & Latin        & 33622  \\
fr      & French               & Latin        & 100337 \\
de      & German               & Latin        & 100265 \\
el      & Greek                & Greek        & 17317  \\
he      & Hebrew               & Hebrew       & 24483  \\
hi      & Hindi                & Devanagari   & 20240  \\
hu      & Hungarian            & Latin        & 31999  \\
is      & Icelandic            & Latin        & 20164  \\
id      & Indonesian           & Latin        & 17297  \\
it      & Italian              & Latin        & 85175  \\
jp      & Japanese             & Japanese     & 98366  \\
ko      & Korean               & Hangul       & 30890  \\
lv      & Latvian              & Latin        & 17247  \\
lt      & Lithuanian           & Latin        & 17232  \\
ml      & Malayalam            & Malayalam    & 19817  \\
mr      & Marathi              & Devanagari   & 20069  \\
ne      & Nepali               & Devanagari   & 20092  \\
nb      & Norwegian            & Latin        & 36811  \\
or      & Oriya                & Oriya        & 19153  \\
pl      & Polish               & Latin        & 34711  \\
pt      & Portuguese           & Latin        & 37229  \\
pa      & Punjabi              & Gurmukhi     & 20026  \\
ro      & Romanian             & Latin        & 17149  \\
ru      & Russian              & Cyrillic     & 20108  \\
sr      & Serbian              & Latin        & 17165  \\
sk      & Slovak               & Latin        & 17255  \\
sl      & Slovenian            & Latin        & 17300  \\
es      & Spanish              & Latin        & 100351 \\
sw      & Swahili              & Latin        & 20170  \\
sv      & Swedish              & Latin        & 36533  \\
ta      & Tamil                & Tamil        & 19807  \\
te      & Telugu               & Telugu       & 19947  \\
th      & Thai                 & Thai         & 17322  \\
tr      & Turkish              & Latin        & 34405  \\
uk      & Ukrainian            & Cyrillic     & 17282  \\
ur      & Urdu                 & Perso-Arabic & 20162  \\
vi      & Vietnamese           & Latin        & 17358  \\
cy      & Welsh                & Latin        & 20207  \\ \bottomrule
\end{tabular}
\caption{Language Distribution in Sphinx Dataset}
\label{tab:lang_table}
\end{table*}

\end{document}